\documentclass{article}

\usepackage{microtype}
\usepackage{graphicx}
\usepackage{booktabs} 
\usepackage{afterpage}

\usepackage{hyperref}

\usepackage[accepted]{icml2025}

\makeatletter
\renewcommand{\Notice@String}{Preprint. 3 Feb 2025.}
\makeatother

\usepackage{amsmath}
\usepackage{amssymb}
\usepackage{mathtools}
\usepackage{amsthm}

\usepackage[capitalize,noabbrev]{cleveref}

\theoremstyle{plain}

\theoremstyle{definition}

\theoremstyle{remark}

\usepackage[textsize=tiny]{todonotes}

\usepackage{float}
\usepackage{booktabs}        
\usepackage{array}           
\newcolumntype{P}[1]{>{\raggedright\arraybackslash}p{#1}}
\usepackage{tabularx}        
\usepackage{threeparttable}  
\usepackage{caption}         
\usepackage{siunitx}         
\usepackage{subcaption}
\usepackage{makecell}
\usepackage{placeins}
\usepackage{listings}
\usepackage{enumitem} 
\usepackage{multirow} 
\usepackage{subcaption}
\DeclareUnicodeCharacter{2032}{'}

\icmltitlerunning{FragmentNet}

\begin{document}

\twocolumn[
\icmltitle{FragmentNet: Adaptive Graph Fragmentation for Graph-to-Sequence Molecular Representation Learning}


\icmlsetsymbol{equal}{*}

\begin{icmlauthorlist}
\icmlauthor{Ankur Samanta}{equal,a}
\icmlauthor{Rohan Gupta}{equal,a}
\icmlauthor{Aditi Misra}{equal,a}
\icmlauthor{Christian McIntosh Clarke}{equal,a}
\icmlauthor{Jayakumar Rajadas}{b}
\end{icmlauthorlist}

\icmlaffiliation{a}{Department of Electrical and Computer Engineering, University of Toronto, Toronto, Canada}
\icmlaffiliation{b}{Advanced Drug Delivery and Regenerative Biomaterials Laboratory, Stanford Cardiovascular Institute, Palo Alto, USA}

\icmlcorrespondingauthor{Ankur Samanta}{ankur.samanta@alumni.utoronto.ca}

\icmlkeywords{Molecular representation learning, graph-to-sequence foundation model, learned graph tokenizer, fragment-based molecule optimization}

\vskip 0.3in
]



\printAffiliationsAndNotice{\icmlEqualContribution} 

\begin{abstract}
Molecular representation learning methods typically tokenize molecules as individual atoms or use rigid, rule-based fragment decompositions, limiting their ability to capture meaningful chemical substructure context. We introduce FragmentNet, a graph-to-sequence model built around a novel adaptive, learned tokenizer that decomposes molecular graphs into chemically valid fragments of adjustable granularity, complemented by chemically aware spatial positional encodings that preserve molecular topology in the resulting sequence. Extending masked pre-training strategies from natural language processing to the molecular domain, we mask and reconstruct molecules at the level of chemically meaningful fragments rather than individual atoms. Evaluating across multiple property prediction benchmarks, we find that pre-training at fragment granularity leads to improved downstream performance on the majority of tasks, demonstrating that tokenization granularity is an important design choice for molecular representation learning.
\end{abstract}

\section{Introduction}
\label{Introduction}

Pre-trained models have transformed natural language processing (NLP) by capturing contextual information from large unlabeled corpora \cite{devlin2018bert}. In parallel, self-supervised learning techniques have been applied to molecular representation learning on graphs of atoms and bonds \cite{gilmer2017neural}. However, early adaptations of Masked Language Modeling at the atom level \cite{hu2019strategies} often fail to capture a broader chemical context, leading to negative transfer where pre-trained models underperform simpler baselines \cite{xia2023molebert}.

Fragment-based modeling has emerged as a more chemically coherent alternative by focusing on substructures rather than individual atoms. However, rule-based methods, such as BRICS \cite{vangala2023pbrics}, can be inflexible and struggle to generalize across chemical space. On the other hand, sequence-based tokenization methods, such as Byte-Pair Encoding on SMILES, often lose topological information.

We introduce FragmentNet, a graph-to-sequence model with three key contributions: (1) a novel adaptive, learned tokenizer that decomposes molecular graphs into chemically valid fragments while preserving connectivity, (2) chemically aware spatial positional encodings that preserve molecular topology when serializing graphs into sequences, and (3) an empirical study of how tokenization granularity interacts with masked pre-training to affect downstream performance. Molecular graphs are tokenized into fragments and compressed using a joint VQVAE+GCN hierarchical encoder, which integrates atom-level features with fragment-level structures. The resulting fragment embeddings, enriched with spatial positional encodings and a molecular descriptor CLS token, are processed by a Transformer trained with Masked Fragment Modeling (MFM).

We evaluate FragmentNet through pre-training with MFM and fine-tuning on the MoleculeNet \cite{wu2018moleculenet} and Malaria \cite{Gamo2010-iq} benchmarks. We show that fragment-level tokenization, when combined with MFM pre-training, outperforms atom-level tokenization on the majority of molecular property prediction tasks, establishing tokenization granularity as a key lever for improving molecular representations. Its fragment-level representations enhance chemical interpretability, as attention maps highlight the substructures driving predictions, and embedding visualizations demonstrate the model’s ability to cluster molecules based on property-specific similarity.

The learned tokenizer also enables a fragment-swapping module for targeted molecular editing, which we demonstrate as a downstream application.

\begin{figure*}[t]
\vskip 0.2in
\begin{center}
\centerline{\includegraphics[width=\textwidth]{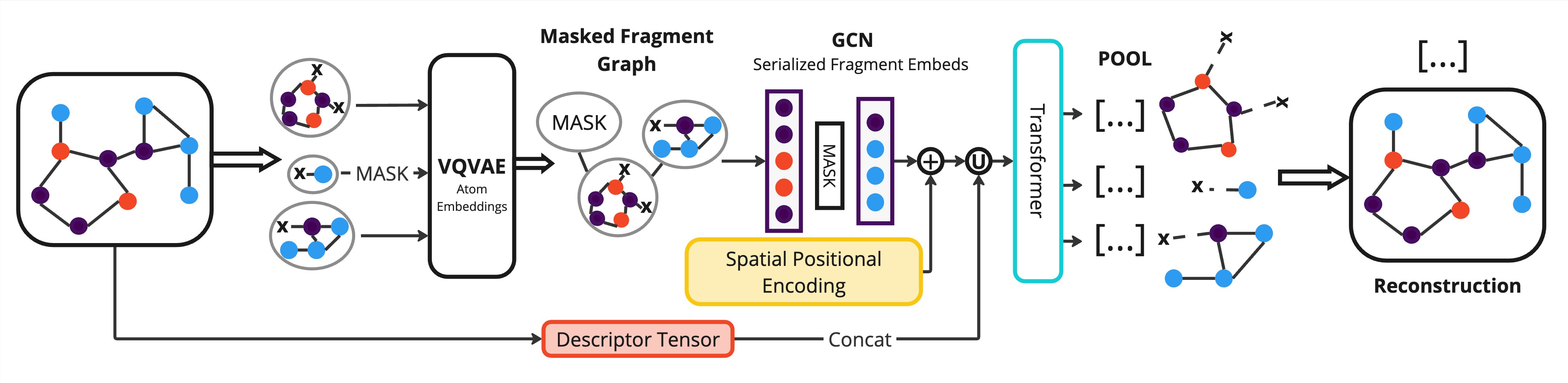}}
    \caption{(Left) A molecule is tokenized into discrete substructures, with one fragment masked. Atom-level attributes are encoded using a VQVAE, and fragment-level attributes are learned using a GCN. The pooled sequence, enriched with spatial positional encodings and a CLS token, is passed through a Transformer. (Right) In MFM, the masked token embedding predicts the masked fragment, while for downstream tasks, the pooled representation encodes the full molecule, which can be reconstructed.}
    \label{fig: system diagram}
    \end{center}
    \vskip -0.2in
\end{figure*}

Our contributions are summarized as follows:
\begin{itemize}
  \item A novel learned adaptive tokenizer for molecular graphs used for Masked Fragment Modeling, producing chemically valid fragments while preserving structural connectivity information.
  \item Novel spatial positional encodings that capture molecular graph topology in a sequence-compatible format, enabling graph-to-sequence modeling.
  \item Empirical demonstration that tokenization granularity is a critical design choice: fragment-level tokenization with MFM pre-training outperforms atom-level tokenization on the majority of tested benchmarks using scaffold splitting.
\end{itemize}

\section{Preliminaries}
\label{Preliminaries}

\subsection{Molecule Models Pre-trained with Masked Atom Modeling on Graphs}
Traditional masked modeling strategies often focus on atom-level masking, limiting a model's ability to capture chemical substructure context. For example, AttrMask \cite{xia2023molebert} randomly hides node and edge attributes in molecular graphs, while MoleBERT \cite{xia2023molebert} employs masked atom modeling (MAM) with a Vector Quantized Variational Autoencoder (VQ-VAE) to enhance atom-level representations. However, atom-level masking often creates broken or chemically inconsistent environments in the visible context, restricting the model's understanding of bonding and functional group interactions. SimSGT \cite{liu2024rethinking} takes a step beyond simple atom masking by training a GNN to predict subgraphs representing the local neighborhoods around masked atoms. Although it offers flexibility in identifying subgraphs, it does not explicitly model interactions between them, limiting its ability to capture long-range dependencies and interactions among functional groups.

\subsection{Masked Fragment Modeling (MFM)}
Masking entire fragments addresses the aforementioned limitation by preserving chemically meaningful contexts, enabling more stable learning of bonding rules and functional relationships \cite{jin2018junction, schwaller2019molecular}. Rooted in organic chemistry principles, this approach preserves contextual dependencies within fragments. It is analogous to reconstructing a multi-word phrase in natural language models, where additional context enhances semantic understanding \cite{sennrich2016neural}. Two primary methods have been explored for its implementation: adaptive tokenization of SMILES strings and rule-based subgraph masking on molecular graphs.

\subsubsection{BERT-Style Pre-Training}
\label{BERT-style pre-training}
Transformer-based architectures \cite{devlin2018bert} have been effective at learning contextual relationships within sequential data. ChemBERTa \cite{chithrananda2020chemberta} extends this paradigm to molecular SMILES representations, using an adaptive tokenizer to split SMILES strings into subsets of characters, leveraging masked language modeling to enhance chemical feature extraction. MolFormer \cite{ross2022large} scales up this technique while incorporating rotary positional embeddings, efficiently pre-training on SMILES sequences from 1.1 billion molecules. MolTRES \cite{park2024moltres} introduces a hierarchical masking strategy for SMILES sequences targeting multiple granularities of chemical substructures, from individual atoms to entire functional groups. However, these SMILES-based transformers often neglect the topological relationships inherent to molecular graphs, sacrificing structurally informed representation learning for modeling efficiency \cite{nguyen2024smiles}.

\subsubsection{Rule-based Masked Fragment Modeling (MFM)}
Existing graph-based methods often rely on rule-based tokenization methods, such as BRICS \cite{vangala2023pbrics}. IBM's r-BRICS extends this by introducing more flexible fragmentation patterns to handle a wider variety of molecules, primarily targeting long aliphatic chains and complex ring structures. UniCorn \cite{feng2024unicorn} employs BRICS-based fragmentation for 2D graph masking. 	However, these methods are limited by their rigidity and might struggle to generalize across diverse chemical spaces \cite{nature2024fragment}. Data-driven approaches overcome this by learning adaptable fragment representations.

\subsection{Graph-Sequence Foundation Models}
While the aforementioned architectures use graph-only or sequence-only representations, combining the two can be beneficial. Graph-BERT \cite{zhang2020graphbert} employs a limited subgraph sampling approach around a single anchor node to transform graphs into sequences. While effective for large knowledge graphs, this method discards too much chemical information, making it less suitable for molecular fragment modeling.

\section{FragmentNet Tokenizing Method}

\subsection{Graph Tokenization}

In natural language processing (NLP), the quality of tokenization is important \cite{vaswani2017attention}, motivating the development of data-driven subword tokenizers \cite{sennrich2016neural,schuster2012japanese}. We introduce a \emph{learned tokenization} technique generalizable to all graph structures. Unlike rule-based methods (e.g., \cite{lewellyn1998recap,degen2008art}), this approach captures empirical substructure distributions directly from data. We explore whether learned tokenization, effective in NLP, can similarly benefit molecular representation learning on graphs.

\section*{Algorithm: Iterative Pairwise Merging of Molecular Fragments}

\begin{algorithm}[tb]
    \caption{Iterative Pairwise Merging of Molecular Fragments}
    \label{alg:pairwise_merging}
\begin{algorithmic}

\STATE \textbf{Input:} Corpus of molecules $\mathcal{C}$, where:
\STATE \hspace{1em} Each molecule $M \in \mathcal{C}$ is represented as a graph with:
\STATE \hspace{2em} Nodes $a_1, a_2, \dots$ indexed by  $\text{atomic\_number}(a)$
\STATE \hspace{2em} Bonds $b_{ij}$ labeled by bond types $\text{type}(b_{ij})$, connecting nodes $a_i$ and $a_j$
\STATE \textbf{Output:} Merged fragments and transformation history

\STATE $\text{merges} \gets \emptyset$
\STATE $\text{next\_node\_ID} \gets \max(\text{atomic\_number}(a) \text{ for all } a \text{ in } \mathcal{C}) + 1$

\FOR{$i = 1$ to $\text{num\_iter}$}
  \STATE $\text{pair\_count} \gets \emptyset$, $\text{node\_count} \gets \emptyset$

  \FOR{each molecule $M \in \mathcal{C}$}
    \FOR{each connected pair of nodes $(a_i, a_j)$ in $M$ with bond $b_{ij}$}
      \STATE $\text{pair\_count}[(a_i, a_j, b_{ij})] \mathrel{+}= 1$
      \STATE $\text{node\_count}[a_i] \mathrel{+}= 1$, $\text{node\_count}[a_j] \mathrel{+}= 1$
    \ENDFOR
  \ENDFOR

  \FOR{each pair $(a_i, a_j, b_{ij})$ in $\text{pair\_count}$}
    \STATE $\text{scores}[(a_i, a_j, b_{ij})] \gets \frac{\text{pair\_count}[(a_i, a_j, b_{ij})]}{\sqrt{\text{node\_count}[a_i] \cdot \text{node\_count}[a_j]}}$
  \ENDFOR

  \STATE $\text{best\_pair} \gets \arg\max_{(a_i, a_j, b_{ij})} \text{scores}[(a_i, a_j, b_{ij})]$

  \FOR{each molecule $M \in \mathcal{C}$}
    \FOR{each pair $(a_i, a_j)$ in $M$ such that $(a_i, a_j, b_{ij}) = \text{best\_pair}$}
      \STATE Replace $a_i$, $a_j$, and $b_{ij}$ with a new node $a_{\text{new}}$ with $a_\text{new}.ID = \text{next\_node\_ID}$
      \STATE $\text{next\_node\_ID} \mathrel{+}= 1$
    \ENDFOR
  \ENDFOR

  \STATE $\text{merges} \gets \text{merges} \cup \{\text{best\_pair}\}$
\ENDFOR
\end{algorithmic}
\end{algorithm}

\subsubsection{Levels of Granularity}

The tokenizer first splits a molecule into its individual atoms and iteratively merges them based on the learned merge history. The number of merges performed controls the size of the tokens. After training the tokenizer for \( T \) iterations, the set of stored merges \( \mathcal{M} \) has length \( T \). A molecule can then be tokenized using the first \( t \) merges from \( \mathcal{M} \) for any \( t \leq T \) without requiring retraining. This flexibility, unique to our tokenizer, allows for adjustable granularity per task.

In this work, we chose 100 merge iterations based on empirical evaluation; further research will explore the impact of the levels of granularity on task-specific downstream performance. \cref{fig: token size distribution} illustrates the impact of these training iterations on the distribution of fragment sizes within our representative dataset of 2 million SMILES strings.

\subsubsection{Token Dictionary}

We tokenized 17,000 molecules to create a vocabulary for our foundation model, then stored the tokens efficiently via hashing. We preserved SMILES, graph representations for reconstruction, and special tokens like [UNK], as discussed further in \cref{app: tokenizer discussion}.

\subsubsection{Treatment of dangling bonds}   
Fragments are created by breaking bonds in the original molecule; we cache the fragments in the token dict, attaching a 'dummy atom' represented by an atom with an atomic number of 0 and no chemical attributes to the other end of the broken bond. This leads to the number and types of broken bonds being an additional differentiator of otherwise identical fragments: A carbon atom with 1 broken single bond,  a carbon atom with 2 broken single bonds, and a carbon atom with 1 broken double bond are all treated as different fragments in the token dictionary.

\subsubsection{Molecule Hashing Method}

To uniquely learn embeddings for different molecular fragments, a one-to-one hashing mechanism is required to distinguish between stereoisomers and tautomers and accommodate dangling bonds. Traditional approaches like SMILES lack uniqueness, as a single fragment can correspond to multiple SMILES strings and do not distinguish isomeric molecules \cite{gilmer2017neural}.

We address this with the Weisfeiler-Lehman (WL) graph hashing algorithm \cite{weisfeiler1968reduction}, ensuring that non-isomorphic graphs receive distinct hashes. Formally, the WL algorithm iteratively refines node labels based on their neighborhood structures:
\begin{equation*}
    l_i^{(t)} = \text{Hash}\left( l_i^{(t-1)}, \{ l_j^{(t-1)} \mid v_j \in \mathcal{N}(v_i) \} \right)
    \label{eq:wl_hash}
\end{equation*}
Here, \( l_i^{(t)} \) is the label of node \( v_i \) at iteration \( t \), and \( \mathcal{N}(v_i) \) is the set of neighbors of \( v_i \).

Atom labels incorporate the atomic number \( Z \), hybridization state \( H \), number of radical electrons \( R \), and hydrogen count \( H_d \). Bond labels include bond type \( B_t \), conjugation status \( C \), stereoisomeric properties \( S \), and ring membership \( R_m \). The resulting hashes construct the token dictionary, mapping molecular fragments to their corresponding learned embeddings.

\section{FragmentNet Model Architecture}
A high-level schematic of FragmentNet's hybrid graph-to-sequence architecture can be found in \cref{fig: system diagram}.

\subsection{Input Representation}
Given a starting molecular graph, FragmentNet receives the following input data: the tokenized fragment graphs (with atom/bond chemical features), the arrangement of the fragments relative to each other in the molecule, and the fragment charges \cref{app: fragment charges}.

\subsection{VQVAE-GCN Encoder for Hierarchical Input Embeddings}

We propose a VQVAE-GCN encoder to construct hierarchical molecular fragment embeddings by integrating Vector Quantized Variational Autoencoders (VQ-VAEs) and Graph Convolutional Networks (GCNs). VQ-VAEs encode discrete atomic-level features \cite{van2017neural}, mapping atomic features into a discrete latent space. For an input \( \mathbf{x} \), the encoder \( E \) maps it to \( \mathbf{z}_e \), which is quantized to the closest vector \( \mathbf{z}_q = \arg\min_{\mathbf{c} \in \mathcal{C}} \|\mathbf{z}_e - \mathbf{c}\|_2 \) in a learned codebook \( \mathcal{C} \). The decoder reconstructs \( \hat{\mathbf{x}} = D(\mathbf{z}_q) \), and the training loss is:
\[
\mathcal{L}_{\text{VQ-VAE}} = \|\mathbf{x} - \hat{\mathbf{x}}\|_2^2 
+ \|\text{sg}[\mathbf{z}_e] - \mathbf{c}\|_2^2 
+ \beta \|\mathbf{z}_e - \text{sg}[\mathbf{c}]\|_2^2
\]
where \( \beta \) balances the commitment loss, and \( \text{sg}[\cdot] \) denotes stop-gradient. Atomic embeddings are averaged to form fragment-level representations, \( \mathbf{z}_{\text{VQ}, f} = \frac{1}{|f|} \sum_{i \in f} \mathbf{z}_{\text{VQ}, i} \).

The GCN captures fragment-level relationships by aggregating features from neighboring nodes within the discrete fragments. Node embeddings are combined in graph convolutions using global mean pooling to generate compressed fragment-level feature representations. The final fragment embedding is obtained by combining VQ-VAE and GCN features as \( \mathbf{r}_f = \mathbf{T}(\mathbf{z}_{\text{VQ}, f}) + \mathbf{z}_{\text{GCN}, f} \), where \( \mathbf{T} \) is a learnable transformation. This integration leverages the strengths of both architectures: VQ-VAEs encode discrete atomic features, while GCNs model structural relationships between atoms. 

\subsection{Graph Spatial Positional Encodings (SPEs)}

After computing embeddings for each fragment, we embed spatial information about the fragments' arrangement in the molecule using Graph Spatial Positional Encodings (SPEs) inspired by \cite{zhang2020graphbert}. We employ three types of SPEs: Hop-based Positional Encoding, Weisfeiler-Lehman (WL) Absolute Positional Encoding, and Coulomb Matrix Positional Encoding. 

\textbf{Hop-based Positional Encoding} captures a node's relative 'connectedness' by aggregating its hop distances from all other nodes in the graph. For a molecular graph \( G = (V, E) \), the hop distance \( H_{i,j} \) from node \( v_i \) to node \( v_j \) is defined as the length of the shortest path between them, or 0 if no path exists. For each node \( v_i \), we define its hop-based encoding as \( \mathbf{h}_i = \sum_{k=1}^{N} \text{Emb}_{\text{hop}}(H_{i,k}) \), where the sum aggregates the encoding of the hop distance relative to all other nodes \( v_j \).

\textbf{Weisfeiler-Lehman (WL) Absolute Positional Encoding} labels nodes uniquely based on their position within the graph structure, ensuring that nodes in two graphs with identical structures receive the same labels. After \( T \) iterations of label refinement using the Weisfeiler-Lehman algorithm, each node \( v_i \) is assigned a unique WL role ID \( w_i \), which is embedded as \( \mathbf{w}_i = \sum_{j=1}^{N} \text{Emb}_{\text{WL}}(w_j) \). Appendix~\ref{app: wl hashing} demonstrates how isomeric molecules receive distinct WL-based hashes through this approach despite sharing the same molecular formula.

\textbf{Coulomb Matrix Positional Encoding} models molecular fragment interactions based on inverse-square law distances. Given node charges \( q_i \) and a fixed distance \( d_0 \), the mean Coulomb interaction for node \( v_i \), aggregated over all anchor nodes, is embedded as \( \mathbf{c}_i = \sum_{j=1}^{N} \text{Emb}_{\text{clb}}(C_{i,j}) \), where \( C_{i,j} \) is defined as:
\[
C_{i,j} = \frac{1}{N} \sum_{k=1}^{N} \left(0.5\, Z_j^{2.4} \delta_{j,k} + \frac{Z_j Z_k}{d_0^2} \left(1 - \delta_{j,k}\right)\right)
\]
Here, \( \delta_{j,k} \) represents the Kronecker delta. This formulation ensures that Coulomb interactions are considered relative to all anchor nodes by directly incorporating both the self-interaction term \(0.5\, Z_j^{2.4} \delta_{j,k}\) and the pairwise interaction term \(\frac{Z_j Z_k}{d_0^2} \left(1 - \delta_{j,k}\right)\) within the aggregate equation.

\subsubsection{Combined Positional Encoding} This encoding integrates hop-based, WL, and Coulomb positional encodings by summing \( \mathbf{h}_i \), \( \mathbf{w}_i \), and \( \mathbf{c}_i \) for each fragment-node \( \mathbf{v}_f \). The aggregated encoding \( \mathbf{p}_f \) comprehensively encapsulates spatial and topological information, contributing to the fragment embedding \( \mathbf{r}_f + \mathbf{p}_f \). Further details and visualizations are provided in \cref{app: spatial posenc visualization}.

\subsection{Molecular Descriptor CLS Token}
In transformer-based architectures, the CLS token aggregates the overall context of the input sequence for classification tasks \cite{devlin2018bert}. In FragmentNet, we replace this token with a Molecular Descriptor Vector, prepending it to the sequence of fragment tokens. The molecular descriptor vector uses RDKit's \texttt{CalcMolDescriptors} method; we compute a vector \( \mathbf{d} = [d_1, d_2, \dots, d_M] \) of standard molecular descriptors. This vector, denoted \( \mathbf{d}_{\text{CLS}} \), is refined during training by attention layers to encode holistic molecular features for downstream tasks.

\subsection{Transformer Encoder Layers}

With the graph structures serialized into a sequence of fragment embeddings, we pass them through a series of BERT transformer encoder blocks \cite{devlin2018bert}, consisting of multi-head self-attention and feed-forward networks with residual connections and layer normalization. The series of attention layers models fragment-to-fragment relationships and captures contextual dependencies and interactions critical for effective structural representation learning.

\subsection{Property Prediction Head}

The property prediction head processes the fragment network's sequence output using max pooling across the sequence length, chosen for its effectiveness in preserving salient features while maintaining padding invariance \cite{suarez2018evaluation}. The pooled representation is then linearly projected to match the target property dimension.

\subsection{Fragment Swapping Module}
\label{frag:swap}
We propose a fragment-swapping algorithm that systematically replaces target fragments with alternatives while preserving the chemical integrity of the core molecular scaffold. The granularity of fragments can be dynamically adapted to suit specific tasks, enabling optimizations such as tailoring molecular analogues for pharmacokinetics or structure-activity relationships (SAR). This serialization process bypasses the need for substructure matching or computationally expensive graph search methods typically used in traditional fragment-based workflows \cite{vangala2023pbrics}. \cref{fragment swapping algorithm} discusses the algorithm and implementation details.

\section{Training Method}

\subsection{Masked Fragment Modeling Pre-training}
\label{sec:mfm}
We pre-train using Masked Fragment Modeling (MFM). The molecular graph is serialized into a sequence of chemically valid fragments, preserving the structural information needed to generate spatial positional encodings (SPE). The SPE encodes the original graph relationships within this sequence. We denote the serialized molecular graph as \( \mathbf{F} = [f_1, f_2, \dots, f_N] \), where each \( f_i \)represents a fragment in the sequence.

In the MFM task, we randomly mask a single fragment \(f_i\) from \( \mathcal{M} \subseteq \{1, 2, \dots, N\} \) and replace it with a special [MASK] token, resulting in the masked sequence \( \mathbf{F}_{\text{masked}} \). A key benefit of our graph serialization is that the token can be directly masked in the sequence, ensuring no data leakage from the original masked fragment to subsequent layers. The model is then trained to predict the original fragment at the masked positions using the context provided by the unmasked fragments. The loss function for MFM is defined as $
\mathcal{L}_{\text{MFM}} 
= - \log P\bigl(f_i \mid \mathbf{F}_{\text{masked}}\bigr)$.

Based on our analysis of fragment sequence length distributions in \cref{app: frag token size dist}, we limit masking to a single token per sequence to preserve context. Prior research indicates diminishing returns with high masking rates in small models \cite{wettig2023mask}. This conservative approach balances task difficulty and model capacity, with plans to explore percentage-based masking in future work.

\subsection{Property Prediction Evaluation}

We evaluated our model's performance on molecular property prediction tasks using datasets from MoleculeNet \cite{wu2018moleculenet} and Malaria \cite{Gamo2010-iq} . These datasets cover physical chemistry, biophysics, and physiology properties, providing a comprehensive benchmark. We employed scaffold splitting with an 80-10-10 train-validation-test ratio. The training was conducted with 10 random seeds for each dataset, and the results were aggregated.

\subsection{Compute and Training Configuration}
Due to funding and compute limitations, FragmentNet was configured and trained on a MacBook Pro M2 laptop using Apple’s Metal Performance Shaders framework. For details on the training setup, methods, hyperparameters, and compute optimization, see \cref{app:model_training_configuration}.

\section{Results and Discussion} 
We compare fragment-level and atom-level tokenization, each with and without MFM pre-training, evaluated using scaffold splitting on MoleculeNet \cite{wu2018moleculenet} and Malaria \cite{Gamo2010-iq} benchmarks. FragmentNet-Atom (15.2M parameters) and FragmentNet-Fragment (17.4M parameters) are both pre-trained on 2M molecules. Blank entries indicate that the model did not produce results for specific datasets.

\begin{table*}[ht]
\vskip 0.15in
\begin{center}
\caption{Performance of FragmentNet Variants on Property Prediction Datasets from MoleculeNet \cite{wu2018moleculenet} and Malaria \cite{Gamo2010-iq} using scaffold splitting.}
\begin{tablenotes}
\footnotesize
[Y] pre-trained, [N] Not pre-trained. FragmentNet-Atom: num\_iters $=0$. FragmentNet-Fragment: num\_iters $=100$. Best results are \textbf{bold}. Format: Median(SD).
\end{tablenotes}
\label{tab:combined_results}
\vskip 0.15in
\begin{small}
\begin{threeparttable}
\begin{tabularx}{\textwidth}{@{}X *{4}{c} *{3}{c}@{}}
\toprule
\textbf{Model} &
\multicolumn{4}{c}{\textbf{Classification (ROC-AUC) ↑}} &
\multicolumn{3}{c}{\textbf{Regression (RMSE) ↓}} \\
\cmidrule(lr){2-5} \cmidrule(lr){6-8}
 & \textbf{BBBP} & \textbf{Tox21} & \textbf{ToxCast} & \textbf{BACE} & \textbf{ESOL} & \textbf{Lipo} & \textbf{MAL} \\
\midrule
FragmentNet-Atom [N]        & 58.8(1.7) & 65.3(2.5) & 56.4(0.2) & 70.8(1.7) & 1.398(0.005) & 1.093(0.005) & 1.159(0.003) \\
FragmentNet-Fragment [N]    & 60.6(2.0) & 61.0(7.5) & 52.7(0.3) & 66.9(0.9) & 1.429(0.042) & 1.119(0.006) & 1.158(0.001) \\
FragmentNet-Atom [Y]        & 71.2(2.2) & 72.4(0.7) & \textbf{61.2(0.3)} & \textbf{79.8(0.8)} & 1.089(0.058) & 0.860(0.011) & 1.116(0.012) \\
FragmentNet-Fragment [Y]    & \textbf{71.4(1.6)} & \textbf{73.0(1.0)} & 61.1(1.0) & 78.7(1.6) & \textbf{0.999(0.034)} & \textbf{0.835(0.004)} & \textbf{1.092(0.001)} \\
\bottomrule
\end{tabularx}
\end{threeparttable}
\end{small}
\end{center}
\vskip -0.1in
\end{table*}

\textbf{Takeaway 1 - Token Granularity Impacts Task Complexity and Structural Context.} \\
We empirically demonstrate the importance of an adaptive graph tokenizer for chemically accurate masked fragment modeling. Token granularity, controlled by the number of merge iterations, affects the complexity and informativeness of the Masked Fragment Modeling (MFM) task. At 0 merge iterations (smallest granularity), fragments consist of a single atom, its bond information, and dummy atoms indicating neighboring connections. When pre-trained on 2 million molecules, as shown in (\cref{app: pre-training curves}), this setup leads to stable convergence, achieving good reconstruction accuracy within the first epoch. Increasing granularity through additional merges (100 iterations) makes the task more complex and chemically informative, with slower but sustained learning, highlighting the potential for improved performance when paired with larger, more diverse datasets. Larger tokens capture higher-level structural motifs, enabling better generalization in downstream tasks. Our pre-trained models consistently outperform un-pretrained models across benchmarks, verifying the effectiveness of MFM and addressing the negative transfer issues common in atom-level masking.

\textbf{Takeaway 2 - Token Granularity Can Be Optimized for Task-Specific Performance.} \\
Our results suggest that token granularity, which controls the size and distribution of fragment tokens, can be treated as an optimizable hyperparameter to improve both pre-training and downstream task performance. As shown in \cref{tab:combined_results}, without pre-training, neither fine-grained nor coarse-grained tokens consistently outperform the other across all tasks. However, with pre-training, fragment-level tokenization shows a clear advantage on the majority of benchmarks, suggesting that the interaction between granularity and pre-training is key. By adjusting the number of merge iterations and token dictionary distribution, future work could explore optimal granularity for specific chemical properties or molecular tasks.

\textbf{Takeaway 3 - Fragment Tokenization Outperforms Atom Tokenization with Pre-training.} \\
With MFM pre-training, fragment-level tokenization outperforms atom-level tokenization on 5 of 7 datasets (\cref{tab:combined_results}), with the remaining two being near-ties. This advantage does not hold without pre-training, where atom-level tokenization performs better on most tasks. This suggests that the benefits of coarser tokenization are unlocked specifically through pre-training, which learns meaningful fragment-level representations that transfer effectively to downstream tasks.

\textbf{Takeaway 4 - Fragment Tokenization Enables Chemically Informed Pre-training.} \\
By compressing molecular graphs into fewer, chemically meaningful tokens, fragment-based tokenization shortens sequence lengths and lowers the computational cost typical of Transformer-based models. Our approach trains on a modest setup (a single laptop) using only 2 million molecules, yet the pre-trained fragment model shows substantial gains over the un-pretrained variant, suggesting that MFM with adaptive tokenization is a more chemically informed pre-training strategy. 
\subsection{Dataset Segregation Visualizations}
\label{Dataset Segregation Visualizations}
\begin{figure}[h]
\vskip 0.2in
\begin{center}
\includegraphics[width=1.0\linewidth]{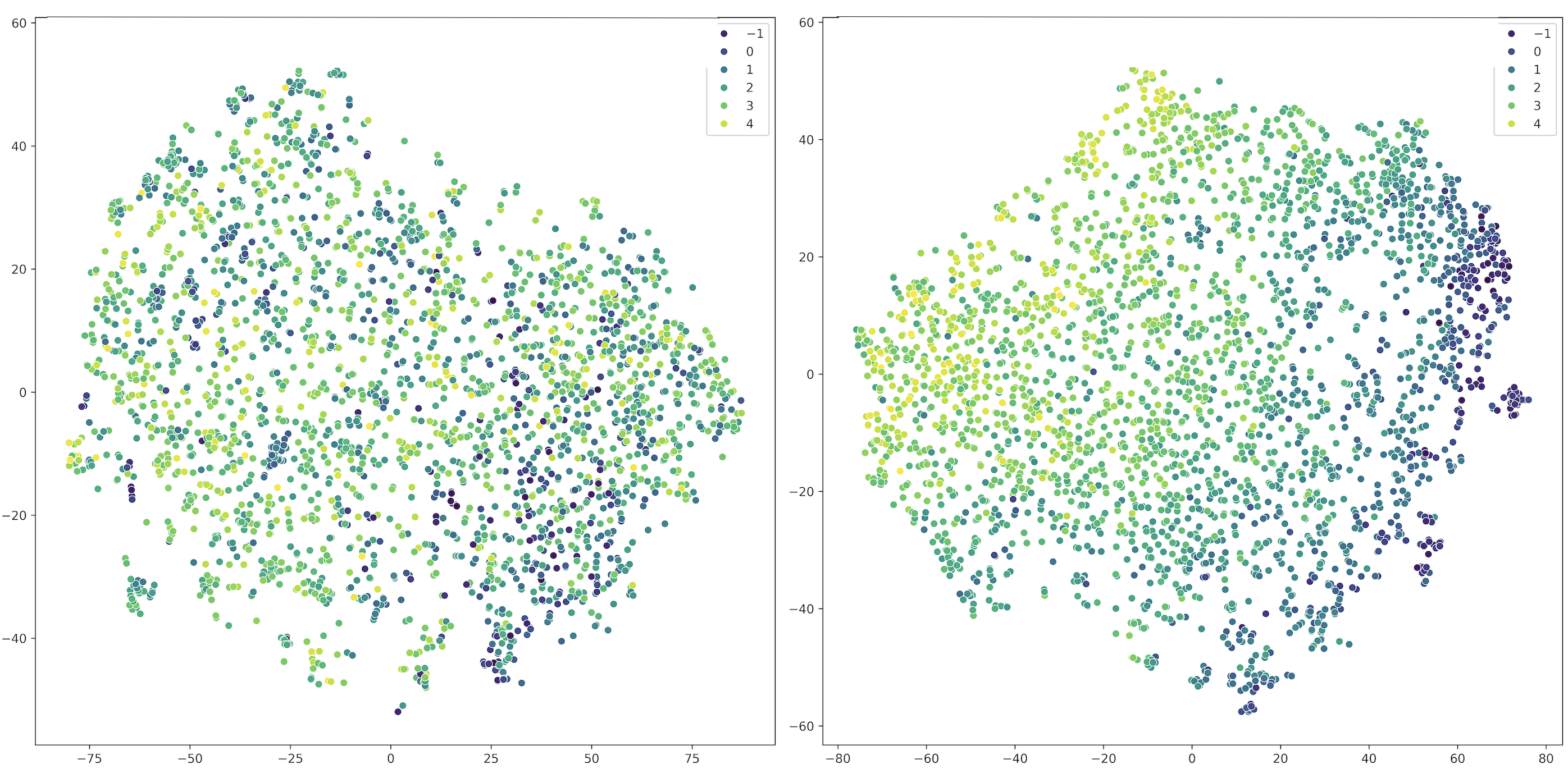}
    \caption{t-SNE of Embedding space for Lipophilicity prediction task before (left) and after (right) fine-tuning, extracted from the second-to-last model layer. Color depicts true values of each data-point.}
    \label{fig:Lipo t-sne}
    \end{center}
    \vskip -0.2in
\end{figure}

We leverage t-SNE to visualize how the embedding space evolves during fine-tuning. Taking the Lipophilicity dataset as an example, we analyze embeddings from the penultimate layer of the fine-tuned model. In \cref{fig:Lipo t-sne}, before training, the points corresponding to molecules with different logD values are randomly scattered across the embedding space. Through training, they become increasingly organized, with emerging clusters that suggest meaningful separations related to molecular properties, including lipophilicity. t-SNE effectively captures global patterns and clustering within the molecular dataset, offering insights into overall relational trends. However, it is less suited for detailed chemical interpretability at the level of individual molecules. Additional t-SNE visualizations are provided in \cref{t-sne appendix analysis}.

\subsection{Attention Maps}
\begin{figure}
\vskip 0.2in
\begin{center}
\centerline{\includegraphics[width=0.5\textwidth]{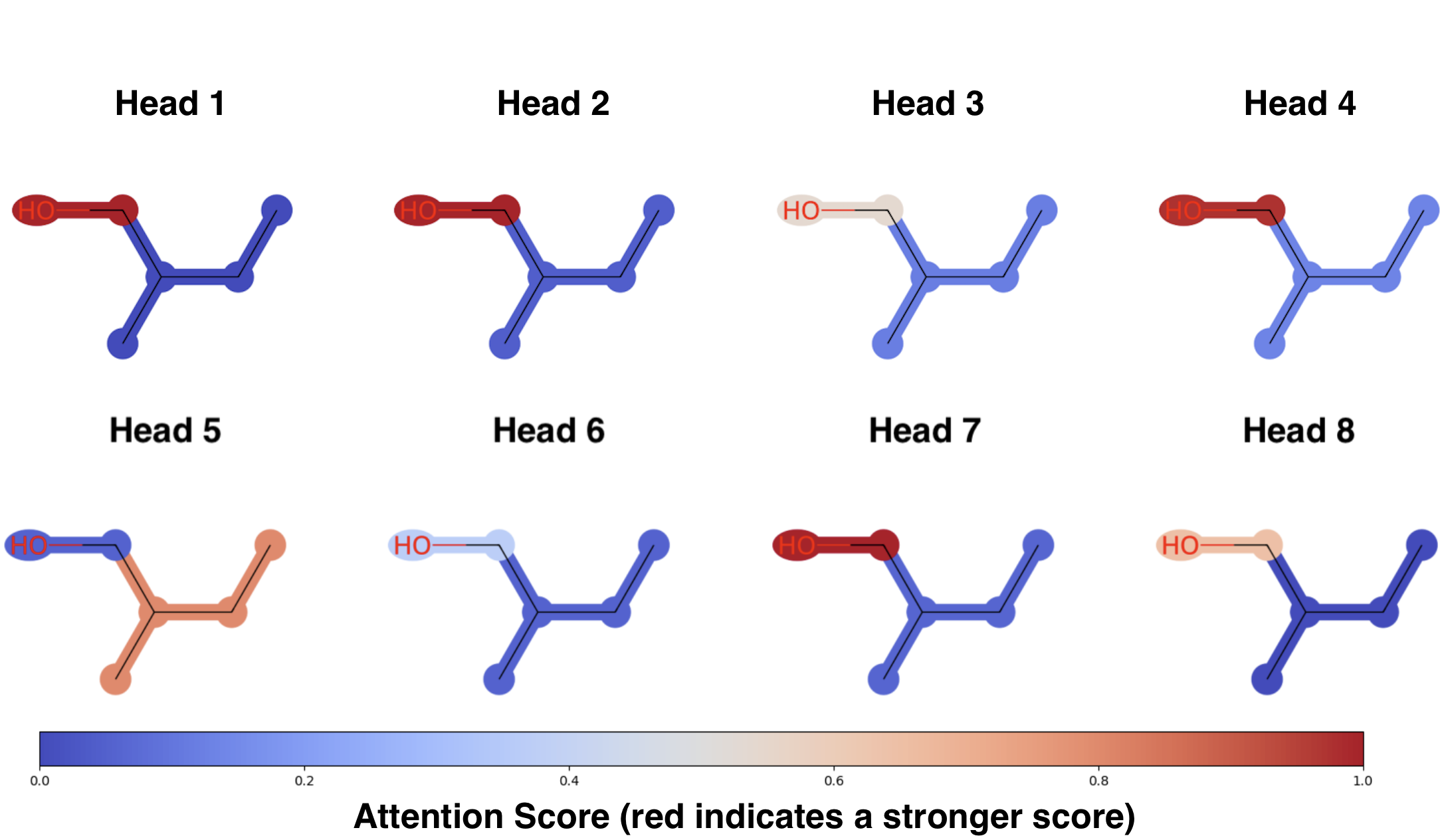}}
    \caption{2-methyl-1-butanol, CCC(C)CO Attention Map for ESOL (aqueous solubility)}
    \label{fig:2-methyl-1-butanol att map}
\end{center}
\vskip -0.2in
\end{figure}
Our approach reveals chemically intuitive patterns by capturing fragment-fragment relationships. As shown in \cref{fig:2-methyl-1-butanol att map}, different attention heads naturally identify distinct molecular features that influence solubility. Most heads emphasize the hydroxyl group (-OH) (red regions), aligning with its role in forming hydrogen bonds with water. Notably, one head focuses on the hydrophobic alkyl fragments—the methyl substituent and carbon backbone—which limit water solubility. This demonstrates how attention mechanisms capture the competing effects of hydrophilic and hydrophobic fragments, highlighting chemical interpretability. Additional attention-map analyses are provided in \cref{Appendix: attention maps discussion}.

\subsection{Fragment Swapping Discussion}
To illustrate the capabilities of the fragment swapping module discussed in \cref{frag:swap}, we start with the molecule Ibuprofen. The adaptive tokenizer serializes Ibuprofen into tokenized fragments, allowing us to identify and modify the carboxylic acid and methyl groups precisely. We then swap out fragments at the target modification site for alternatives using our fragment-swapping algorithm, which avoids the need for substructure matching or graph-based searches.

\begin{figure}[h]
\vskip 0.2in
\begin{center}
\centerline{\includegraphics[width=0.5\textwidth]{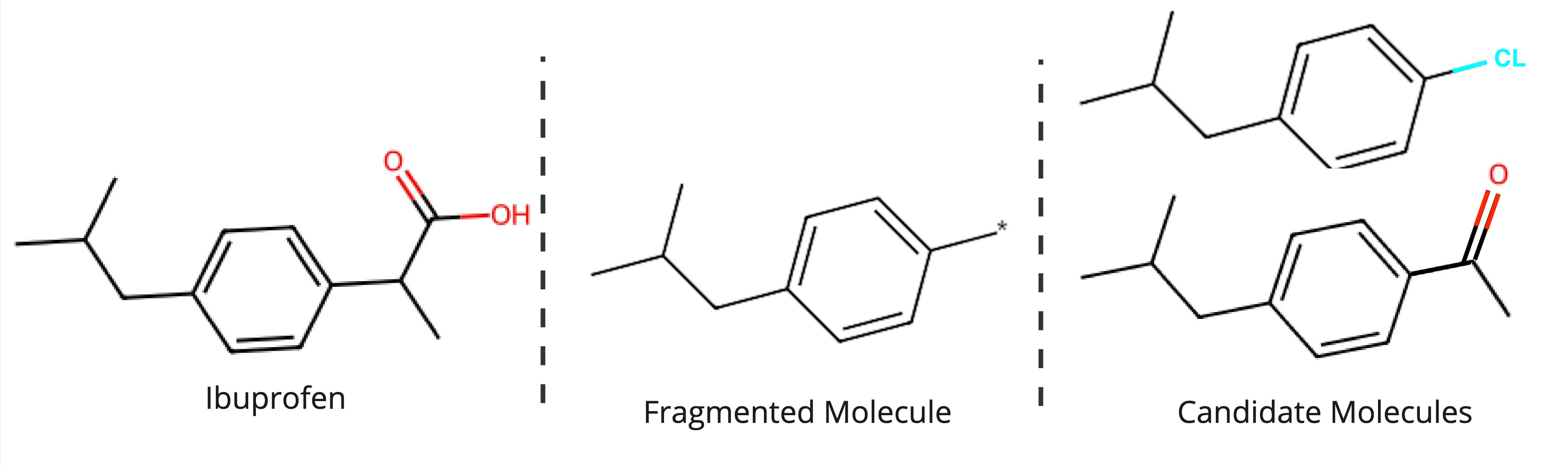}}
    \caption{Left: Ibuprofen. Center: We fragment off the carboxylic acid and methyl groups. Right: We generate two alternative molecules through fragment swapping, CC(C)Cc1ccc(Cl)cc1, and CC(C)Cc1ccc(cc1)C(=O)O}
    \label{fig:ibuprofen fragment swapping}
    \end{center}
    \vskip -0.2in

\end{figure}

Using this approach, we generate two chemically valid analogues: \texttt{CC(C)Cc1ccc(Cl)cc1} and \texttt{CC(C)Cc1ccc(cc1)C(=O)O} (\cref{fig:ibuprofen fragment swapping}). These analogues preserve the molecular scaffold of Ibuprofen while introducing structural diversity, demonstrating how the module facilitates efficient, precise analogue generation.

By leveraging adaptive tokenization to simplify and serialize molecular representation, we can rapidly and systematically explore chemical spaces. This is particularly valuable for applications such as drug discovery and materials design. Further examples are in \cref{appendix:generate_analogues}.

\section{Conclusion, Limitations, and Future Work}
We introduce FragmentNet, a graph-to-sequence model built around a novel adaptive, learned graph tokenizer and chemically aware spatial positional encodings that preserve molecular topology when serializing graphs into sequences. We show that pre-training at the granularity of molecular fragments rather than individual atoms leads to improved downstream performance on the majority of benchmarks, establishing tokenization granularity as a critical design choice for molecular representation learning. Due to the compute constraints of this project, with all training conducted on a single laptop, we only evaluate two granularity levels, use a small pre-training dataset of 2 million molecules with a tokenizer vocabulary trained on a 17k subset, mask only a single fragment per sequence, and keep the model small at 15-17M parameters. Future work can investigate the optimal granularity for molecular representation learning, scale to larger models and datasets, and explore how to best design graph-to-sequence architectures and evaluate these tokenization principles in other such architectures.


\bibliography{example_paper}
\bibliographystyle{icml2025}

\newpage
\appendix
\onecolumn

\section{Tokenizer Discussion}
\label{app: tokenizer discussion}
\subsection{Sample Readout from Token Dictionary}

The below token dictionaries were generated using FragmentNet's adaptive tokenizer with 0 and 100 merge iterations for atoms and fragments, respectively, rather than relying on rule-based fragmentation approaches. Each entry in the token dictionary is a tuple comprising the fragment's hash ID, its SMILES representation (where \texttt{*} denotes a placeholder atom), a graph representation that includes node features (\texttt{x}), edge connections (\texttt{edge\_index}), edge features (\texttt{edge\_attr}), and the number of atoms (\texttt{num\_atoms}), along with additional metadata (currently \texttt{None}).

\subsection{Granularity \texttt{num\_iters=0} Sample Entries}

Within our tokenization framework, the smallest level of granularity retains the structure of common fragments through bonds and dummy atoms; it keeps a single atom within the fragment. However, retaining the connection points makes the fragment more informative than a singular atom. For example:

1. Double-bonded oxygen (=O) with one connection point:
\begin{lstlisting}
(('a55e82208e74da72a2918a7c63ed516a', '*=O', Data(x=[2, 133], edge_index=[2, 2], edge_attr=[2, 147], num_atoms=2), None))
\end{lstlisting}
Functional groups like aldehydes (*CH=O), ketones (C()=O), or carboxylic acids (*C(=O)OH) could have been preserved if fragments were tokenized at a coarser granularity.

2. Carbon with one double bond and two connection points:
\begin{lstlisting}
(('4bbdcb3d627f9a34eeb0c15cfb26b630', '*C=*', Data(x=[3, 133], edge_index=[2, 4], edge_attr=[4, 147], num_atoms=3), None))
\end{lstlisting}
Coarser fragments might better represent substructures like vinyl groups (CH=CH2), acrylic esters (CH=C()O), or conjugated dienes (CH=CH-CH=CH).

3. Nitrogen with three connection points:
\begin{lstlisting}
(('99b11f1d4f8f8cb2956189b2117d5b02', '*N(*)*', Data(x=[4, 133], edge_index=[2, 6], edge_attr=[6, 147], num_atoms=4), None))
\end{lstlisting}
Such fragments could include amide groups (N()C(=O)), sulfonamides (N()S(=O)2*), or tertiary amines (N(CH3)2).



\subsection{Fragment Token Dictionary Sample Entries}

As we increase the granularity, fragment-based tokenization creates more chemically meaningful tokens. Note that the fragment dictionary retains the atom fragments, which can be used as needed. Below are a few samples from the fragment token dictionary:

1. A chiral cyclic amine with a carbonyl group:
\begin{lstlisting}
(('dd071a4542f22e18743e1be19dcd20b2', '*[C@@H]1CCCN1C=O', Data(x=[8, 133], edge_index=[2, 16], edge_attr=[16, 147], num_atoms=8), None))
\end{lstlisting}

2. A hydroxyl group attached to a benzene ring with a placeholder connection point:
\begin{lstlisting}
(('ec766bc887f489cc1b0d875eb133c2d6', '*Oc1cccc(*)c1*', Data(x=[10, 133], edge_index=[2, 20], edge_attr=[20, 147], num_atoms=10), None))
\end{lstlisting}




\subsection{Token Dictionary Length}
As this is an adaptive tokenizer, the size of the token dict tends to scale based on the size of the token dict training dataset and the tokenizer granularity.

Due to our computing constraints, it was not feasible for us to train a token dictionary with the same dataset as our pre-training task, which consisted of 2 million SMILES. Therefore, we decided to use a 17k subset of the pre-training dataset and saw an atom token dictionary of length 98 and a fragment token dictionary length of 8737.

\subsection{Fragment Token Size Distribution}
\label{app: frag token size dist}

\begin{figure}[H]
\vskip 0.2in

\begin{center}
    \centerline{\includegraphics[width=0.45\textwidth]{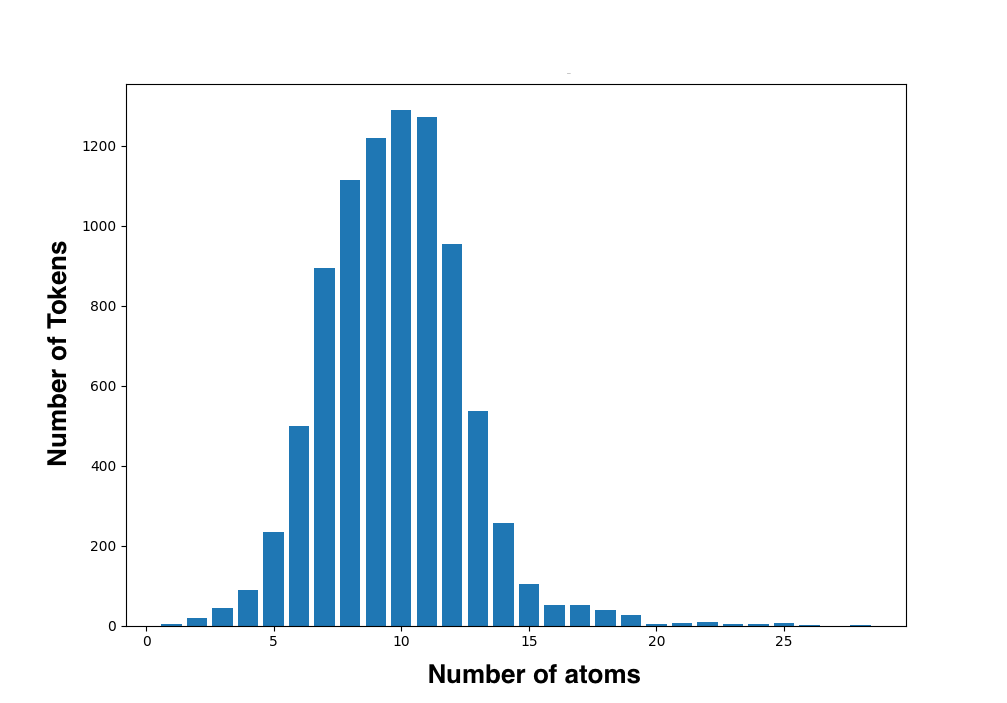}}
    \centerline{\includegraphics[width=0.45\textwidth]{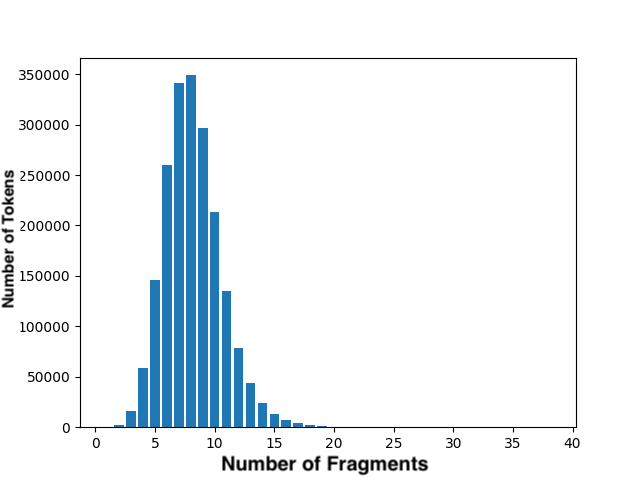}}
    \caption{(Top) Distribution of number of atoms in each token in our token dictionary, (Bottom) Number of fragments for each molecule in the pre-training dataset}
    \label{fig: token size distribution}
\end{center}
    \vskip -0.2in

\end{figure}

\cref{fig: token size distribution} illustrates key statistics about our molecular tokenization scheme after applying 100 merge operations. The first distribution shows the number of fragments per molecule, indicating that most molecules contain approximately 7 fragments on average. However, the distribution exhibits a tail of larger molecules with more fragments. This suggests that our tokenizer effectively compresses molecular structures while maintaining meaningful substructures.

The second distribution displays the number of atoms per token, revealing that most tokens correspond to fragment sizes of around 10 atoms. This indicates that our tokenization approach balances fragmentation and meaningful chemical subunits, relevant for learning molecular representations.

In our Masked Fragment Modeling (MFM) task, we leverage this tokenization scheme by randomly masking a fragment and training the model to predict the missing part. This task encourages the model to capture local chemical dependencies while learning a high-level structural understanding of molecules. The observed fragment distribution suggests that our tokenizer segments molecules into chemically meaningful building blocks.

\subsubsection{Discussion on Tokenizer Granularity}

The atom and fragment token dictionaries demonstrate the flexibility of our tokenizer framework, which allows granularity to be parameterized based on the number of merge iterations. This enables direct comparison of masked pre-training tasks at different granularities within the same modeling framework.

By tuning the number of merge iterations, we can influence the distribution of token granularities learned by the tokenizer. This parameterization enables exploration of how granularity distribution affects different prediction tasks. However, further research is needed to evaluate the optimality of different granularity distributions specific to downstream property task performance.

\section{Weisfeiler-Lehman Molecule Hashing}
\label{app: wl hashing}
The WL-hashing algorithm we have adapted to molecules can uniquely hash them while distinguishing many forms of isomerism, which is not straightforward with other string-based molecular representations such as SMILES \cite{weininger1988smiles} and InChI \cite{heller2015inchi}. In Table 2, we review a well-known case of isomerism and note that molecules that exhibit forms of isomerism are assigned unique hashes through our proposed WL-molecular hashing methods, despite having the same smiles strings.
\begin{table}[H]
\centering
\begin{tabular}{m{0.1\textwidth}|m{0.4\textwidth}|m{0.4\textwidth}}
\textbf{Isomerism Type} & \textbf{Molecule 1} & \textbf{Molecule 2} \\ 
\midrule
Geometric Isomerism & \includegraphics[width=0.2\textwidth]{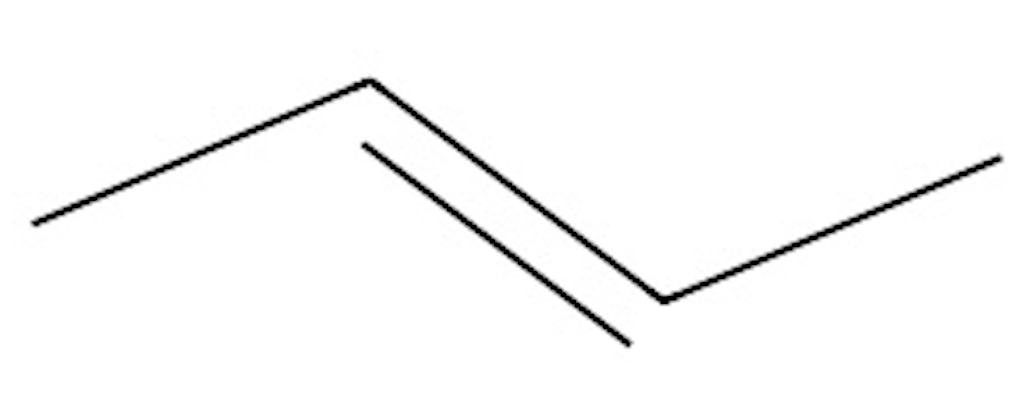} \newline \textit{Cis-2-butene. Smiles: C/C=C/C \newline Hash: c8cf1d7c17272855dc186976e7075a43} & \includegraphics[width=0.2\textwidth]{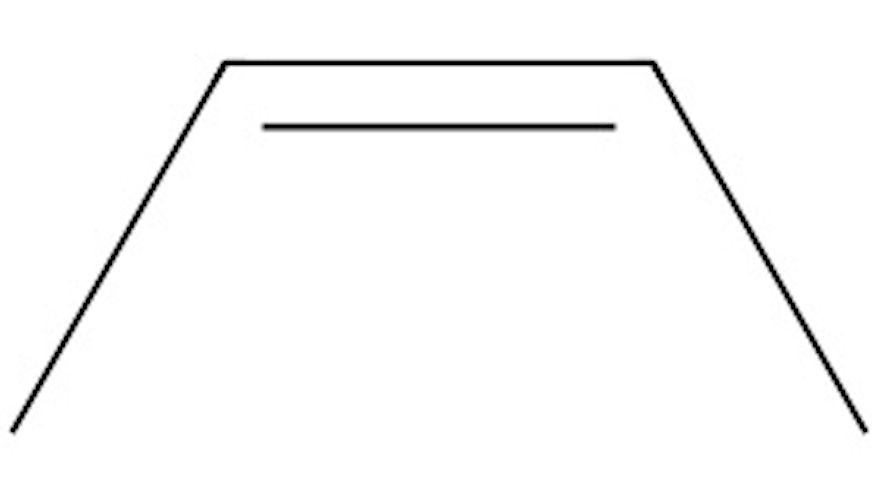} \newline \textit{Trans-2-butene. Smiles: C/C=C\textbackslash C \newline Hash: 610a093419128912a6f09f82b9321df3} \\ 
\label{tab:isomerism table}
\end{tabular}
\caption{Common forms of isomerism in molecules.}
\end{table}



\section{Fragment Charges}
\label{app: fragment charges}
\textbf{Fragment Charges.} We define the charge of a fragment as the sum of the Gasteiger-Marsili partial charges for all non-dummy atoms in the fragment, i.e., \( Q_{\mathcal{F}} = \sum_{a \in \mathcal{F}} q_a \), where \( q_a \) is the partial charge of atom \( a \) if the total charge is not a number (NaN), set \( Q_{\mathcal{F}} = 0 \).

\section{Spatial Positional Encodings Visualization}
\label{app: spatial posenc visualization}
\begin{figure} [H]
\vskip 0.2in
\begin{center}
    \centerline{\includegraphics[width=0.6\textwidth]{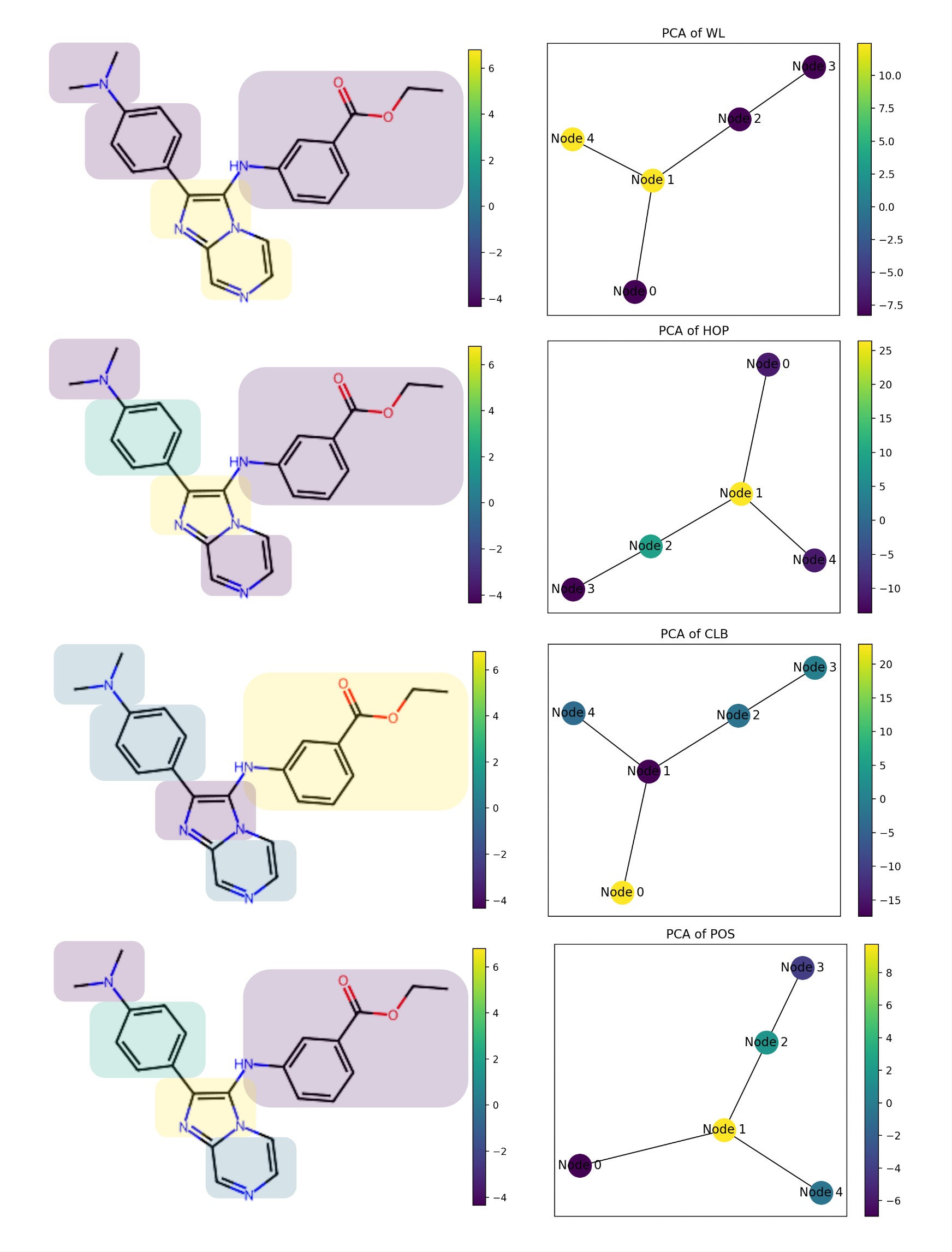}}
    \caption{The principal component differentiation of the molecule-fragments under Weisfeiler-Lehman (WL) positional-encoding, multi-anchor hop-based (HOP), Coulomb (CLB) and the aggregated encoding (POS).}
    \label{fig: posenc visualization}
    \end{center}
\vskip -0.2in

\end{figure}

Figure \cref{fig: posenc visualization} Provides a visualization of the principal component analysis (PCA) applied to fragments of a molecule under different positional encoding schemes: Weisfeiler-Lehman (WL), multi-anchor hop-based (HOP), Coulomb (CLB), and their aggregated encoding (POS). The WL encoding captures the relative positional relationships between nodes by using a breadth-first search (BFS)-type algorithm to encode the connectedness of nodes within the molecular graph. The HOP encoding generalizes positional relationships further by considering the connectivity of each node to all other nodes in the graph, providing a more global sense of structural embedding. The CLB encoding, in contrast, leverages electrochemical properties, encoding nodes based on the Coulombic interactions, which emphasize the physical and chemical characteristics of the molecule. When aggregated, these encodings (POS) yield a comprehensive differentiation of molecular fragments, as seen in the rightmost PCA plots. This demonstrates these encodings' complementary strengths in capturing structural and chemical nuances, effectively distinguishing molecular substructures.

\section{Fragment Swapping Algorithm}
\label{appendix:generate_analogues}
The \textit{Fragment Swapping Algorithm} is designed to generate chemically valid analogues of a given molecule by systematically substituting molecular fragments at predefined positions. The input molecule ($\mathcal{M}$) contains dummy atoms denoting points of substitution, and a set of candidate fragments ($f_i \in \mathcal{F}$) with matching dummy atoms is provided. The algorithm checks chemical validity by matching the bond environments of the dummy atoms in both the input molecule and candidate fragments, followed by reconstruction and sanitization of the substituted structure.

This approach facilitates the controlled exploration of structural diversity, maintaining the core scaffold of the original molecule while varying its substituents. This is particularly valuable for drug discovery and materials design, where analogues with similar structural frameworks but differing properties can offer insights into structure-activity relationships (SAR).

\subsection{Algorithm}
\label{fragment swapping algorithm}
The pseudocode for the \textit{Fragment Swapping Algorithm} is presented below:

\begin{algorithm}[h]
\caption{Generate Analogues}
\begin{algorithmic}
\STATE \textbf{Input:} Molecule $\mathcal{M}$ with dummy atoms denoting dangling bonds for fragment insertion.
\STATE Set of candidate fragments $f_i \in \mathcal{F}$ with dummy atoms denoting dangling bonds.
\STATE \textbf{Output:} List of chemically valid generated analogues, with identical structure to inputted molecule $\mathcal{M}$ but with fragment substitution.

\STATE \textbf{function} \texttt{get\_atom\_bond\_hash}($\mathcal{M}$) 
    \STATE \hspace{1em} hash dict $\gets \emptyset$
    \STATE \hspace{1em} \textbf{for} dummy atom $a_i$ \textbf{in} $\mathcal{M}$ \textbf{do}
        \STATE \hspace{2em} num bonds per type $\gets \emptyset$
        \STATE \hspace{2em} \textbf{for} bond $b_{i,j}$ \textbf{connected to} $a_i$ \textbf{do}
            \STATE \hspace{3em} num bonds per type[type($b_{i,j}$)] $\gets$ num bonds per type[type($b_{i,j}$)] + 1
        \STATE \hspace{2em} \textbf{end for}
        \STATE \hspace{2em} hash dict[hash(num bonds per type)].append($a_i$)
    \STATE \hspace{1em} \textbf{end for}
    \STATE \hspace{1em} \textbf{return} hash dict
\STATE \textbf{end function}

\STATE \textbf{for} each $f_i$ \textbf{in} $\mathcal{F}$ \textbf{do}
    \STATE \hspace{1em} fragment hash dict $\gets$ \texttt{get\_atom\_bond\_hash}($f_i$)
    \STATE \hspace{1em} \textbf{if} keys(fragment hash dict) $\neq$ keys(mol hash dict) \textbf{then}
        \STATE \hspace{2em} \textbf{continue}
    \STATE \hspace{1em} \textbf{end if}
    \STATE \hspace{1em} \textbf{if} length(fragment hash dict[$k$]) $\neq$ length(mol hash dict[$k$]) $\forall k \in$ keys(fragment hash dict) \textbf{then}
        \STATE \hspace{2em} \textbf{continue}
    \STATE \hspace{1em} \textbf{end if}

    \STATE \hspace{1em} \textbf{for} key \textbf{in} keys(fragment hash dict) \textbf{do}
        \STATE \hspace{2em} Generate all possible mappings between fragment hash dict[key] and mol hash dict[key].
    \STATE \hspace{1em} \textbf{end for}
    \STATE \hspace{1em} \textbf{for} all unique mappings of dummy atoms \textbf{do}
        \STATE \hspace{2em} For each mapping, reconstruct $\mathcal{A}$ by removing paired dummy atoms and bonds.
        \STATE \hspace{2em} \textbf{if} \texttt{RdKit.Chem.SanitizeMol}($\mathcal{A}$) without error \textbf{then}
            \STATE \hspace{3em} analogues $\gets$ analogues $\cup \mathcal{A}$
        \STATE \hspace{2em} \textbf{end if}
    \STATE \hspace{1em} \textbf{end for}
\STATE \textbf{end for}
\end{algorithmic}
\end{algorithm}

\subsection{Example 1: Aspirin}
\begin{figure}[H]
\vskip 0.2in
\begin{center}
    \centerline{\includegraphics[width=0.75\textwidth]{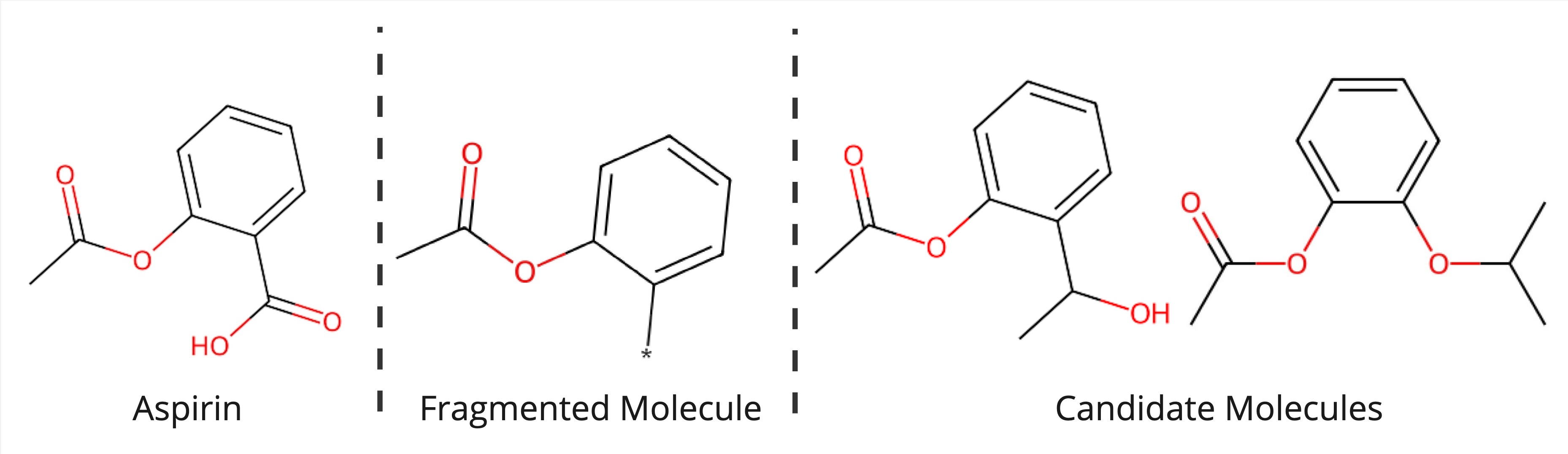}}
    \caption{Fragment swapping example for Aspirin.}
    \label{fig:aspirin_candidates}
    \end{center}
\vskip -0.2in

\end{figure}

In this example, the starting molecule is Aspirin (\texttt{CC(C)CC1=CC=C(C=C1)C(C)C(=O)O}). We fragment off the carboxylic acid functional group and substitute it with two chemically valid fragments. The resulting analogues are depicted in \cref{fig:aspirin_candidates}. 

\textbf{Left:} The original Aspirin molecule. \\
\textbf{Center:} The carboxylic acid group is removed, leaving a dummy atom for substitution. \\
\textbf{Right:} Two analogues are generated: \texttt{CC(=O)Oc1ccccc1C(O)C} and \texttt{CC(=O)Oc1ccccc1OC(C)C}. Both analogues preserve the overall molecular scaffold while introducing new functional groups, potentially altering biological activity or physicochemical properties.

\subsection{Example 2: Diazepam}
\begin{figure}[H]
\vskip 0.2in
\begin{center}
    \centerline{\includegraphics[width=0.75\textwidth]{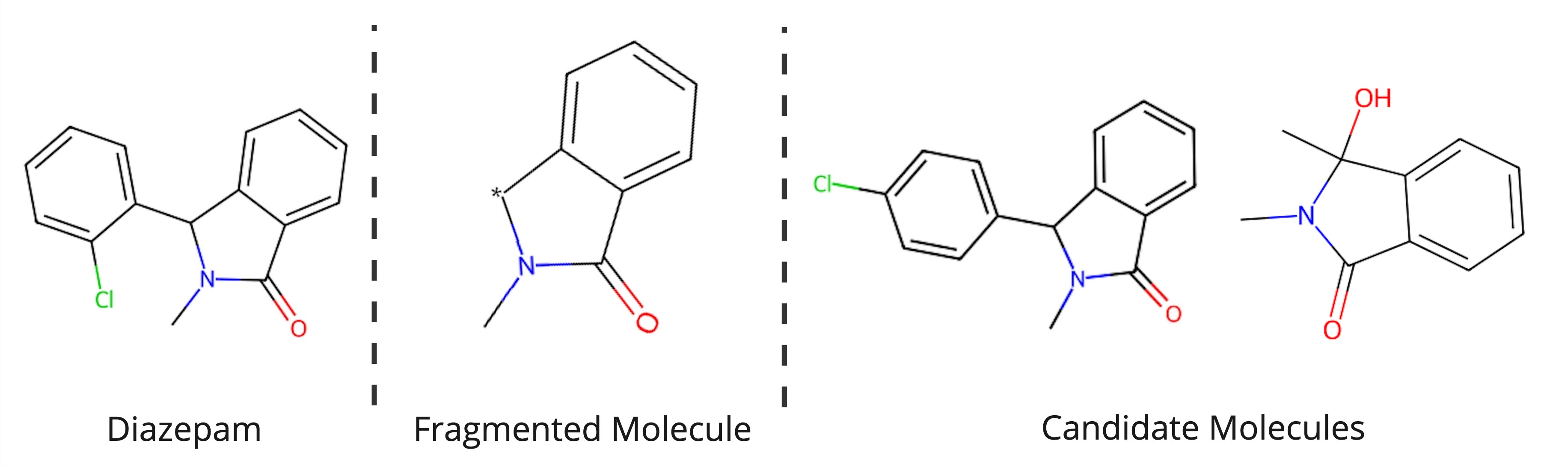}}
    \caption{Fragment swapping example for Diazepam.}
    \label{fig:diazepam_candidates}
    \end{center}
\vskip -0.2in

\end{figure}

In this example, the starting molecule is Diazepam (\texttt{CN1C(=O)c2ccccc2C1c3ccccc3Cl}). We fragment off the chlorobenzene functional group and substitute it with two alternative fragments, as shown in \cref{fig:diazepam_candidates}. 

\textbf{Left:} The original Diazepam molecule. \\
\textbf{Center:} The chlorobenzene group is removed, leaving a dummy atom for substitution. \\
\textbf{Right:} Two analogues are generated: \texttt{CN1C(=O)c2ccccc2C1c1ccc(Cl)cc1} and \texttt{CN1C(=O)C2=CC=CC=C2C1(O)C}. These modifications alter the electronic properties and potential receptor binding of Diazepam derivatives. \\

\subsection{Significance of Adaptive Fragmentation and Fragment Swapping}

Adaptive fragmentation, coupled with the \textit{Fragment Swapping Algorithm}, provides a straightforward way to generate analogues by substituting learned fragments. Future work involves integrating the algorithm with machine learning models to prioritize analogues based on predicted properties.

\section{Property Prediction Embedding Space Visualization}
\label{t-sne appendix analysis}

 Before fine-tuning, embeddings are unstructured across all datasets. Regression tasks like ESOL lack alignment with continuous target values, while classification datasets like BBBP exhibit class overlap.

\begin{figure}
\vskip 0.2in
\begin{center}
    \centerline{\includegraphics[width=0.8\linewidth]{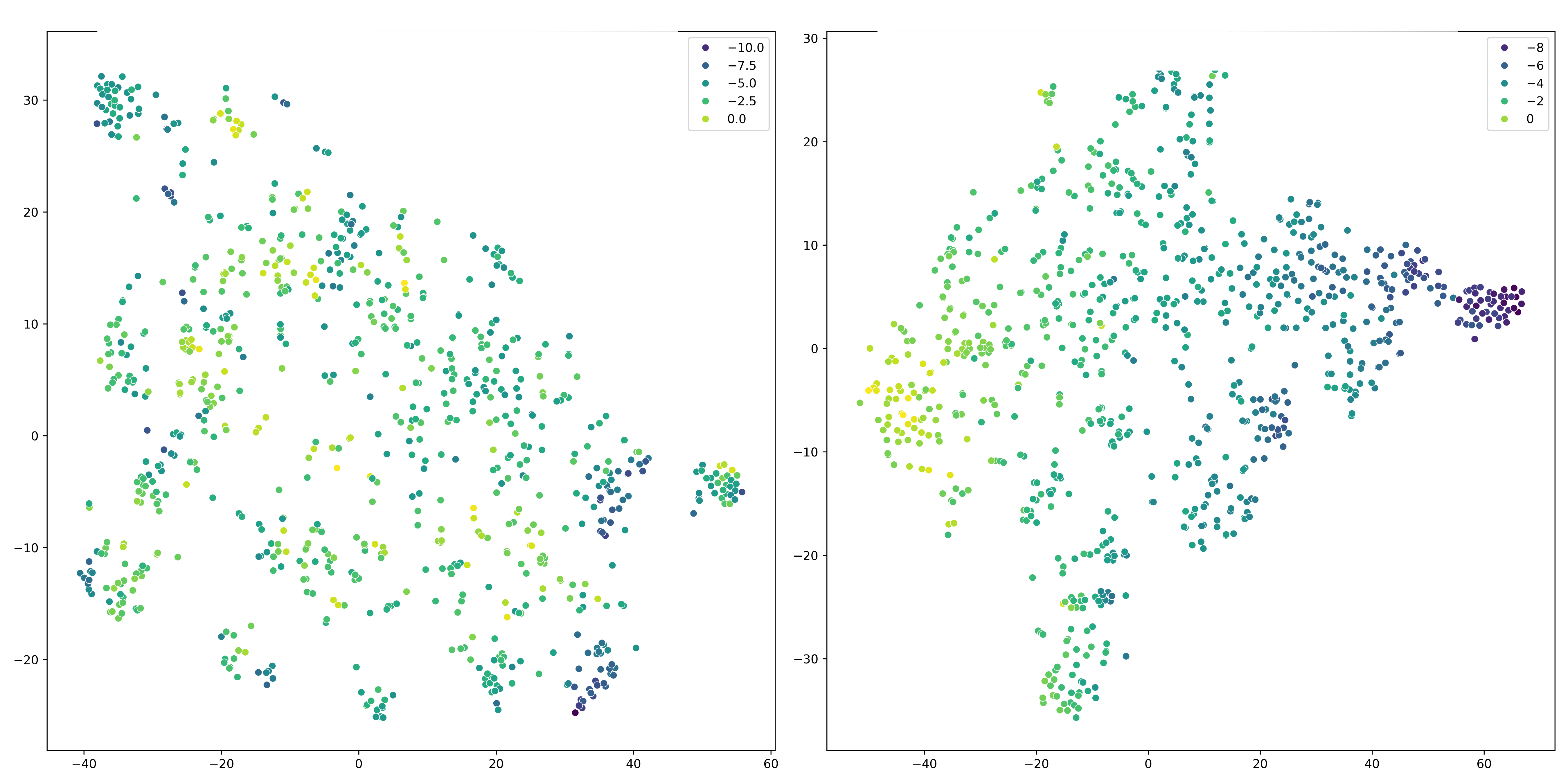}}
    \centerline{\includegraphics[width=0.8\linewidth]{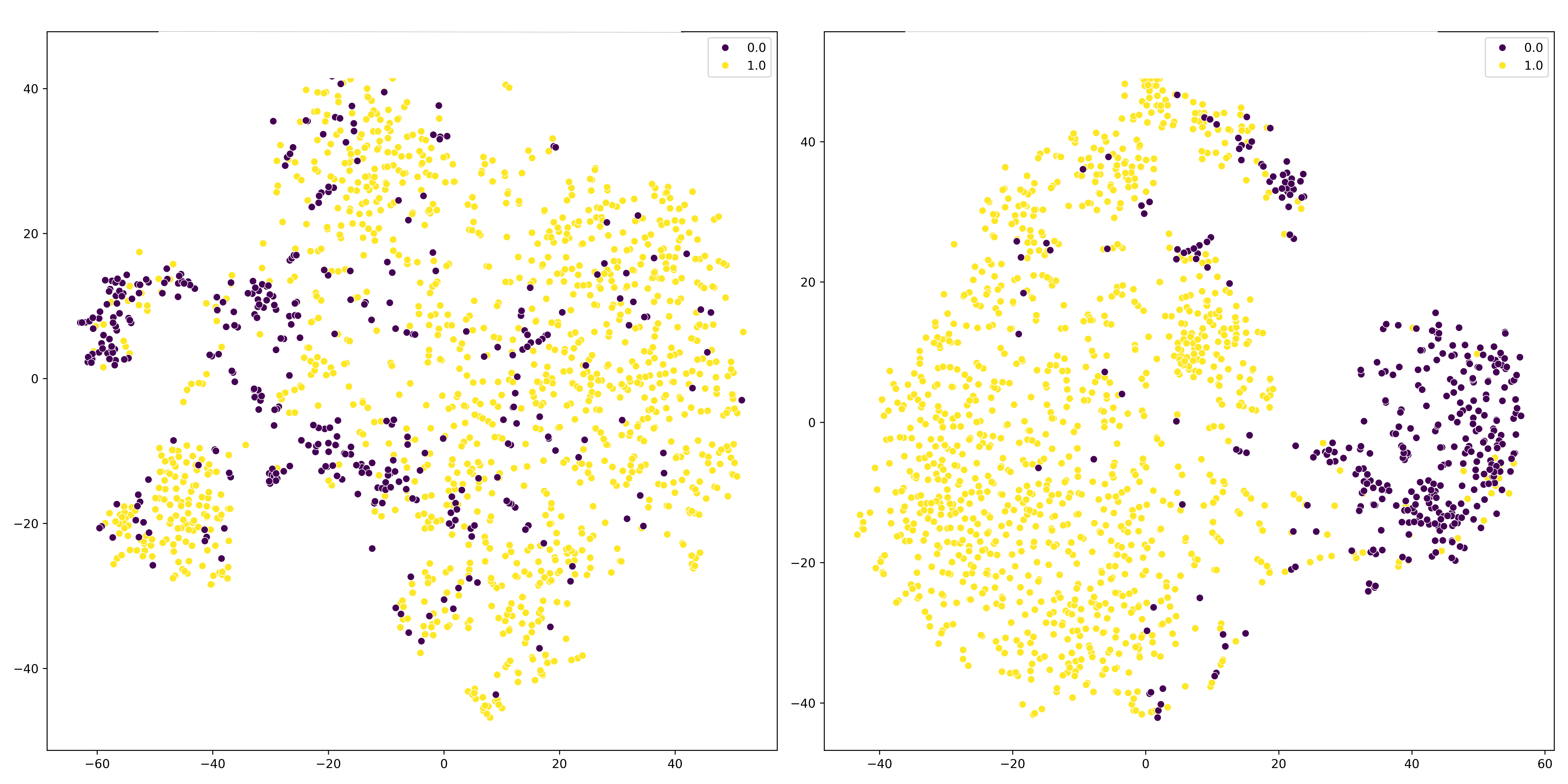}}
    \caption{t-SNE visualization of FragmentNet embedding spaces. Top: ESOL (regression); Bottom: BBBP (classification). Left: before fine-tuning; Right: after fine-tuning. Colors represent true values or labels.}
    \label{fig:app t-SNE}
    \end{center}
\vskip -0.2in
\end{figure}

Fine-tuning improves embedding organization in both regression and classification tasks. In the ESOL dataset containing aqueous solubility data, embeddings initially appear scattered but later align into a gradient reflecting solubility values, indicating improved molecular representation. In contrast, the BBBP dataset, with binary blood-brain barrier permeability labels, initially shows overlapping class embeddings. After fine-tuning, clusters corresponding to labels "0" and "1" emerge, with minor overlap at decision boundaries reflecting the complexity of this classification task. Future work will explore contrastive learning to refine clustering and feature separability further.

\section{Chemical Interpretability of FragmentNet through Attention Maps}
\label{Appendix: attention maps discussion}

We analyze attention maps of two molecules across datasets to showcase the model’s ability to identify chemically meaningful fragments linked to solubility, permeability, and bioactivity. This approach aligns closely with traditional chemical reasoning, improving interpretability for solubility, permeability, and bioactivity predictions.

\subsection{N-methyl-2-amino-3-phenylpropane (BBBP Dataset)}

\begin{figure}[H]
\vskip 0.2in
\begin{center}
    \centerline{\includegraphics[width=0.6\linewidth]{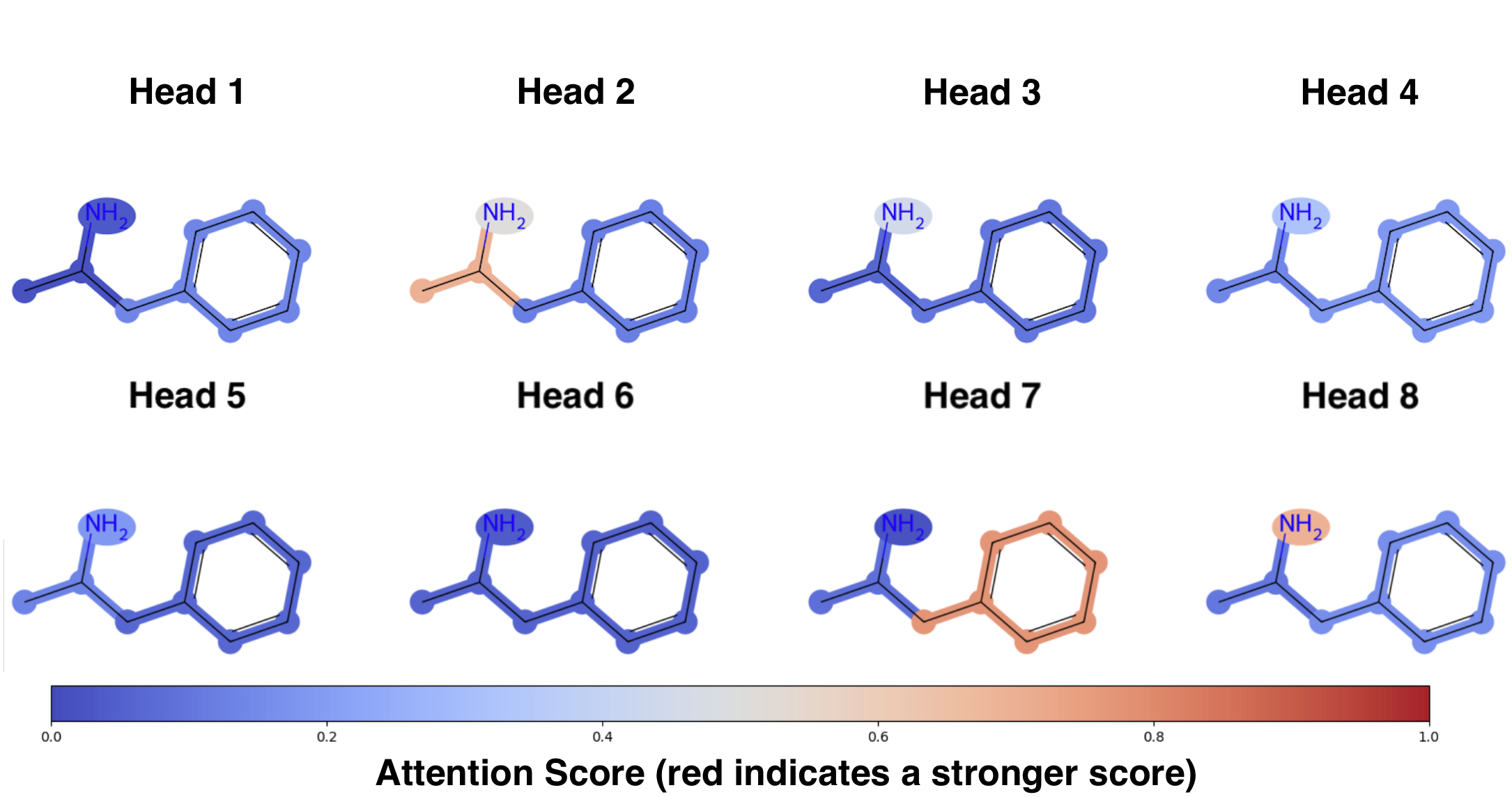}}
    \caption{Attention Map for N-methyl-2-amino-3-phenylpropane (task: brain blood barrier prediction)}
    \label{fig:attention map for N-methyl-2-amino-3-phenylpropane appendix}
\end{center}
\vskip -0.2in

\end{figure}

The compound \textit{N-methyl-2-amino-3-phenylpropane} (\texttt{CC(CC1=CC=CC=C1)N}) from the BBBP dataset shows high blood-brain barrier permeability, and its attention map highlights structural features known to influence BBB transport. The amine group (\(-\text{NH}_2\)) receives strong attention, consistent with its role in hydrogen bonding and potential interactions with transport proteins, while attention to the benzene ring aligns with its contribution to membrane permeability through lipophilic interactions \cite{cornelissen2023explaining}.

\subsection{CC(C)CN(CC1=CC=CC=C1)C2=NC3=CC=CC=C3N=C2 (Malaria Dataset)}

\begin{figure}[H]
\vskip 0.2in
\begin{center}
    \centerline{\includegraphics[width=0.6\linewidth]{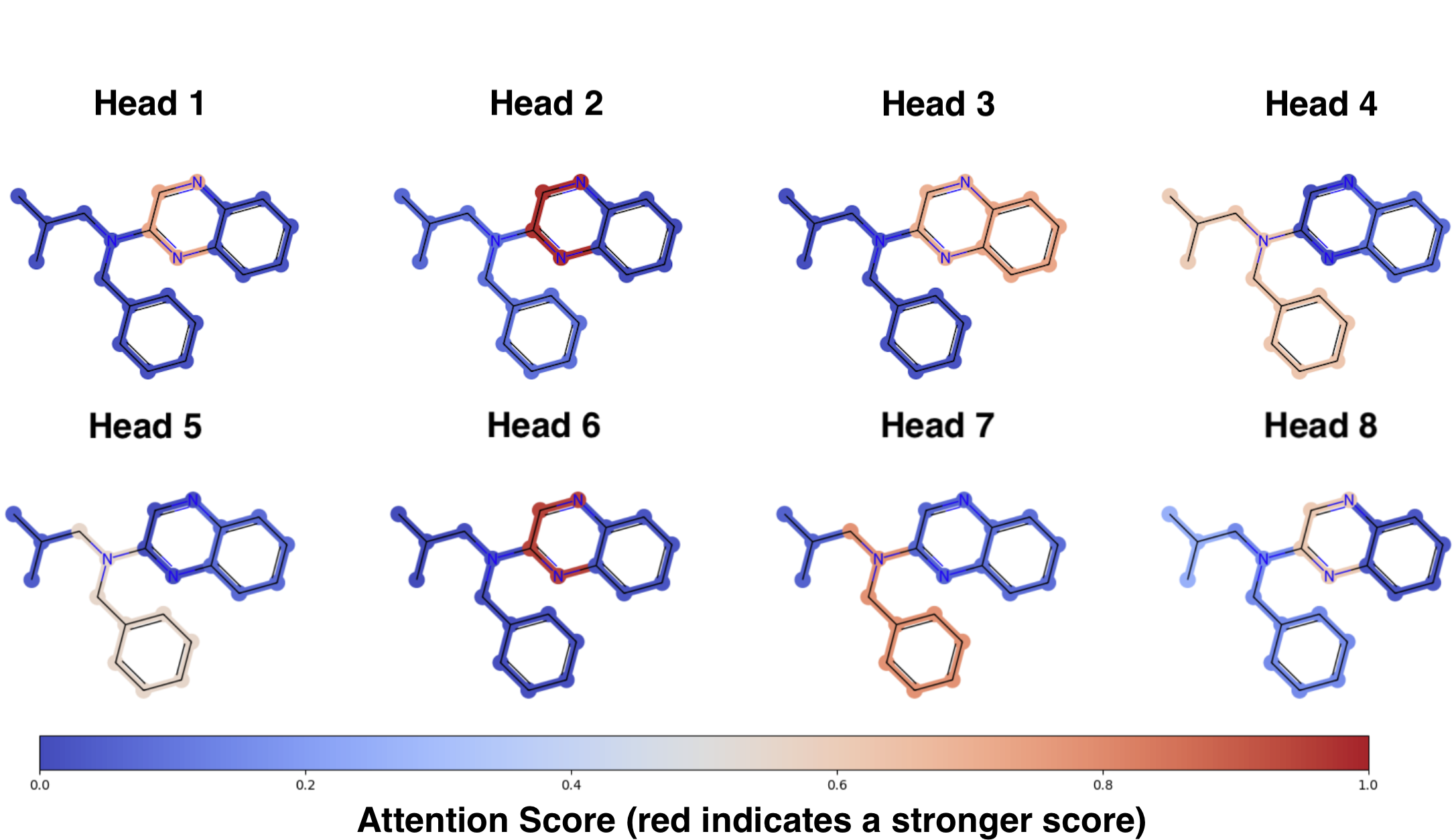}}
    \caption{Attention Map for CC(C)CN(CC1=CC=CC=C1)C2=NC3=CC=CC=C3N=C2 (task: malaria)}
    \label{fig:attention map for CC(C)CN(CC1=CC=CC=C1)C2=NC3=CC=CC=C3N=C2 appendix}
\end{center}
\vskip -0.2in

\end{figure}

For \texttt{CC(C)CN(CC1=CC=CC=C1)C2=NC3=CC=CC=C3N=C2}, the attention map highlights the quinazoline core and benzylamine moiety as key contributors to bioactivity against malaria-related targets. The quinazoline core, a known pharmacophore, facilitates $\pi$-$\pi$ stacking and hydrogen bonding with targets such as dihydrofolate reductase (DHFR), relevant to antimalarial activity \cite{nzila2010preclinical}. The benzyl group enhances binding through hydrophobic interactions, while its flexible linker allows optimal positioning within the binding pocket.

\section{Model and Training Configuration}
\label{app:model_training_configuration}

\subsection{Pre-training Model Configuration}

\begin{table*}[ht]
\centering
\caption{Model and Pre-training Configuration}
\begin{minipage}{0.48\linewidth}
    \centering
    \subcaption{Key Configuration Parameters for FragmentNet}
    \begin{tabular}{@{}lll@{}}
        \toprule
        \textbf{Component} & \textbf{Parameter} & \textbf{Value} \\ \midrule
        \multirow{4}{*}{GCN}  & Node Features  & 133  \\
                              & Feature Dim    & 64   \\
                              & Layers        & 2    \\
                              & Dropout       & 0.2  \\ \midrule
        \multirow{4}{*}{VQVAE} & Codebook Dim  & 133  \\
                               & Dropout       & 0.15 \\
                               & Num Codebooks & 4    \\
                               & Codebook Sizes & [64, 64, 64, 64] \\ \midrule
        Sequence Encoder & Dimension & 256 \\
                         & Layers    & 8   \\
                         & Heads     & 8   \\ \midrule
    \end{tabular}
    \label{tab:magbert_config}
\end{minipage}
\hfill
\begin{minipage}{0.48\linewidth}
    \centering
    \subcaption{Pre-training Hyperparameters}
    \begin{tabular}{@{}ll@{}}
        \toprule
        \textbf{Parameter} & \textbf{Value} \\ \midrule
        Training Set Size      & 2M molecules \\
        Batch Size             & 128 \\
        Optimizer              & Adam \\
        \quad Learning Rate (\texttt{lr}) & \{0.0001, 0.0002, 0.0008\} \\
        \quad Betas (\texttt{betas})       & (0.9, 0.999) \\
        Weight Decay           & 0 \\
        Loss Function          & CrossEntropyLoss \\
        \quad Reduction        & Mean \\ \bottomrule
    \end{tabular}
    \label{tab:hyperparameters_pre-training}
\end{minipage}
\end{table*}

The \textbf{GCN} module captures molecular fragment structure, leveraging two layers with a hidden size of 64 and a dropout rate of 0.2. The \textbf{VQVAE} employs four codebooks of size 64 for vector quantization. The \textbf{Sequence Encoder} and \textbf{BERT} layers share a hidden size of 256 with eight encoder blocks and eight attention heads to ensure effective representation learning. 

The model was pre-trained on a dataset of 2 million molecules using the Adam optimizer with a varying learning rate schedule. The CrossEntropyLoss function guided optimization, with the loss averaged across the elements.

\subsubsection{Fine-tuning Model Configuration}

For fine-tuning, the Adam optimizer maintained a learning rate of 0.0001 with betas set to (0.8, 0.99) and an epsilon of \(1 \times 10^{-8}\) to enhance numerical stability. A batch size of 32 was used, and training was limited to 100 epochs, with early stopping triggered at 40 epochs to mitigate overfitting. A dropout rate of 0.6 and max pooling ensured robust feature selection.

A grid search was conducted over key hyperparameters (\cref{tab:grid_search_hyperparameters}), varying learning rates, batch sizes, hidden dimensions, layers, dropout rates, and early stopping criteria to refine performance further. The best combination was selected based on validation performance across property prediction datasets.

\begin{table*}[hbt!]
\centering
\caption{Fine-tuning Hyperparameters and Grid Search Configurations}
\begin{minipage}{0.48\linewidth}
    \centering
    \subcaption{Fine-tuning Hyperparameters}
    \begin{tabular}{@{}ll@{}}
        \toprule
        \textbf{Parameter} & \textbf{Value} \\ \midrule
        Training Set Size        & Varies by dataset \\
        Batch Size               & 32 \\
        Optimizer                & Adam \\
        \quad Learning Rate (\texttt{lr}) & 0.0001 \\
        \quad Betas (\texttt{betas})       & (0.8, 0.99) \\
        \quad Epsilon (\texttt{eps})       & $1 \times 10^{-8}$ \\
        Number of Epochs         & 100 \\
        Early Stopping Patience  & 40 epochs \\
        Dropout Rate             & 0.6 \\
        Pooling Method           & Max, Mean, Att, CLS (None) \\ \bottomrule
    \end{tabular}
    \label{tab:hyperparameters_fine_tuning}
\end{minipage}
\hfill
\begin{minipage}{0.48\linewidth}
    \centering
    \subcaption{Hyperparameter Grid Search Configurations}
    \begin{tabular}{@{}ll@{}}
        \toprule
        \textbf{Hyperparameter} & \textbf{Values Explored} \\ \midrule
        Learning Rate (\texttt{lr})    & \{0.0001, 0.00008, 0.0002\} \\
        Batch Size                     & \{16, 32, 64\} \\
        Hidden Dimension               & \{64, 128, 256\} \\
        Number of Layers (\texttt{num\_layers}) & \{1, 2, 4\} \\
        Dropout Rate                   & \{0.4, 0.6\} \\
        Early Stopping Patience        & \{10, 40, 60\} \\ \bottomrule
    \end{tabular}
    \label{tab:grid_search_hyperparameters}
\end{minipage}
\end{table*}

\FloatBarrier


\subsection{Training Compute}
All models were trained on a MacBook Pro M2 Max with 32GB of RAM using Apple’s Metal Performance Shaders (MPS) framework. Training on a dataset of 2 million molecules, we completed each epoch in under 11 hours.


\subsection{FragmentNet Pre-training Curves}



\label{app: pre-training curves}
\begin{figure}[H]
\vskip 0.2in
\begin{center}
    \includegraphics[width=0.48\linewidth]{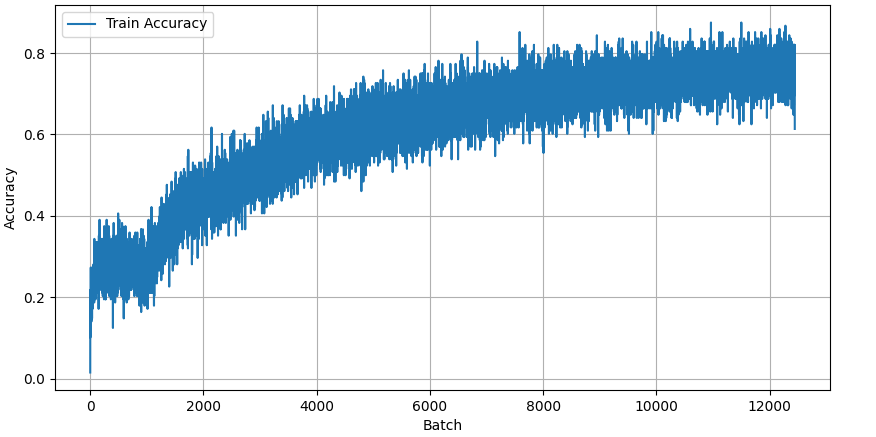}
    \includegraphics[width=0.48\textwidth]{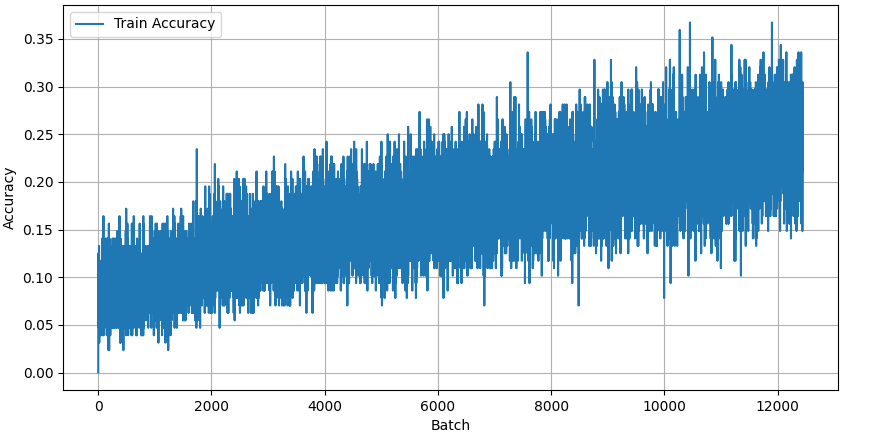}
    \caption{FragmentNet pre-training epoch 1 batch accuracy. Left: \texttt{num\_iters = 0}, Right: \texttt{num\_iters = 100}.}
    \label{fig:pre-training_metrics accuracy}
    \end{center}
\end{figure}

\begin{figure}[H]
\begin{center}
    \includegraphics[width=0.48\linewidth]{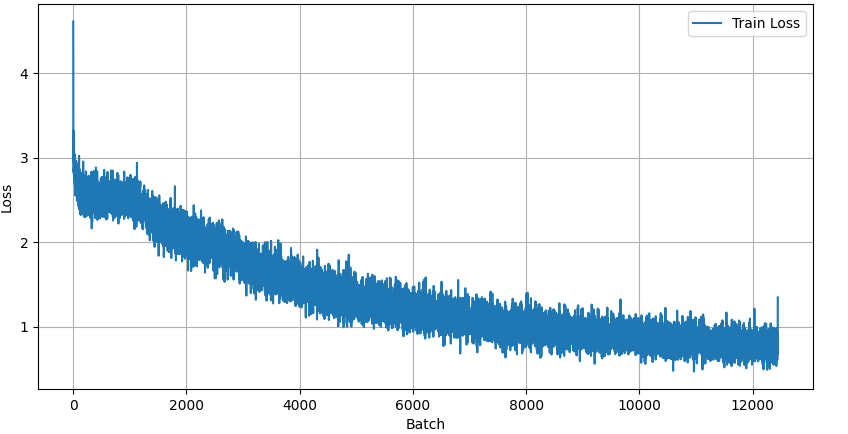}
    \includegraphics[width=0.48\textwidth]{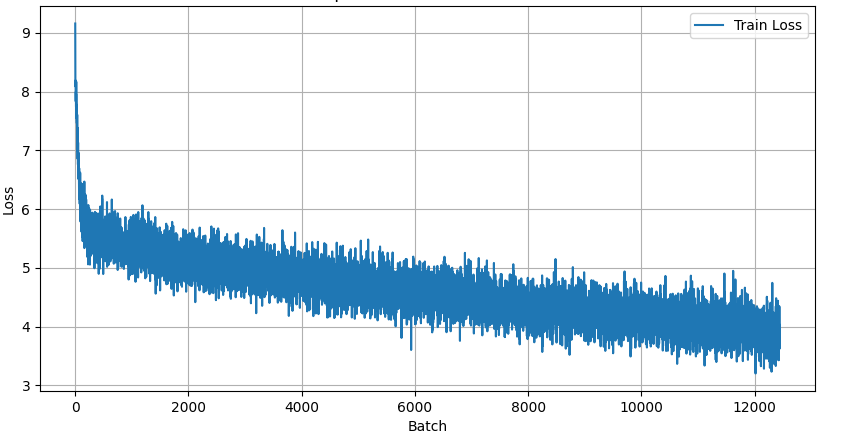}
    \caption{FragmentNet pre-training epoch 1 batch loss. Left: \texttt{num\_iters = 0}, Right: \texttt{num\_iters = 100}.}
    \label{fig:pre-training_metrics loss}
    \end{center}
\vskip -0.2in
\end{figure}

\end{document}